%% file: main.tex
\def\isarxiv{1} %%% for icml submission version, we comment this line

\ifdefined\isarxiv
\documentclass[11pt]{article}

\else
\documentclass{article} % For LaTeX2e
\usepackage{icml2025}

\usepackage{microtype}
\usepackage{hyperref}
\usepackage{url}

\fi

\usepackage{booktabs}
\usepackage{amsmath}
\usepackage{amsthm}
\usepackage{amssymb}
\usepackage{algorithm}
\usepackage{subfig}
\usepackage{graphicx}
\usepackage{grffile}
\usepackage{wrapfig,epsfig}
\usepackage{url}
\usepackage{xcolor}
\usepackage{epstopdf}

\usepackage{bbm}
\usepackage{dsfont}
\usepackage{multirow}
\allowdisplaybreaks

\ifdefined\isarxiv

\usepackage{tikz}
\usepackage{hyperref}  %%% arxiv don't allow this.
\hypersetup{colorlinks=true,citecolor=blue,linkcolor=blue} %%% Zhao : maybe we should comment this in submission.
\usetikzlibrary{arrows}
\usepackage[margin=1in]{geometry}

\else

\fi
\newtheorem{theorem}{Theorem}[section]
\newtheorem{lemma}[theorem]{Lemma}
\newtheorem{definition}[theorem]{Definition}

\newtheorem{proposition}[theorem]{Proposition}

\newtheorem{fact}[theorem]{Fact}
\newtheorem{remark}[theorem]{Remark}
\newtheorem{claim}[theorem]{Claim}

\newcommand{\wt}{\widetilde}

\newcommand{\N}{\mathcal{N}}
\newcommand{\R}{\mathbb{R}}

\renewcommand{\d}{\mathrm{d}}

\DeclareMathOperator*{\E}{{\mathbb{E}}}
\DeclareMathOperator*{\var}{\mathrm{Var}}

\DeclareMathOperator{\poly}{poly}

\DeclareMathOperator{\diag}{diag}

\makeatletter
\newcommand*{\RN}[1]{\expandafter\@slowromancap\romannumeral #1@}
\makeatother

\usepackage{lineno}

\ifdefined\isarxiv
\else
\usepackage[textsize=tiny]{todonotes}

\icmltitlerunning{How Sparse Attention Approximates Exact Attention? Your Attention is Naturally Sparse}
\fi

\begin{document}

\ifdefined\isarxiv

\date{}

\title{How Sparse Attention Approximates Exact Attention?\\Your Attention is Naturally $n^C$-Sparse}
\author{
Yichuan Deng\thanks{\texttt{ycdeng@cs.washington.edu}. The University of Washington.}
\and
Zhao Song\thanks{\texttt{magic.linuxkde@gmail.com}. Simons Institute for the Theory of Computing, University of California, Berkeley.}
\and
Jing Xiong\thanks{\texttt{junexiong@connect.hku.hk}. The University of Hong Kong}
\and
Chiwun Yang\thanks{\texttt{christiannyang37@gmail.com}. Sun Yat-sen University.}
}

\else

% \twocolumn[

% \aistatstitle{How Sparse Attention Approximates Exact Attention? Your Attention is Naturally $n^C$-Sparse}

% \aistatsauthor{ Author 1 \And Author 2 \And  Author 3 }

% \aistatsaddress{ Institution 1 \And  Institution 2 \And Institution 3 } ]

\twocolumn[
\icmltitle{How Sparse Attention Approximates Exact Attention?\\Your Attention is Naturally \texorpdfstring{$n^C$}{}-Sparse}



\icmlsetsymbol{equal}{*}

\begin{icmlauthorlist}
\icmlauthor{Firstname1 Lastname1}{equal,yyy}
\icmlauthor{Firstname2 Lastname2}{equal,yyy,comp}
\icmlauthor{Firstname3 Lastname3}{comp}
\icmlauthor{Firstname4 Lastname4}{sch}
\icmlauthor{Firstname5 Lastname5}{yyy}
\icmlauthor{Firstname6 Lastname6}{sch,yyy,comp}
\icmlauthor{Firstname7 Lastname7}{comp}
\icmlauthor{Firstname8 Lastname8}{sch}
\icmlauthor{Firstname8 Lastname8}{yyy,comp}
\end{icmlauthorlist}

\icmlaffiliation{yyy}{Department of XXX, University of YYY, Location, Country}
\icmlaffiliation{comp}{Company Name, Location, Country}
\icmlaffiliation{sch}{School of ZZZ, Institute of WWW, Location, Country}

\icmlcorrespondingauthor{Firstname1 Lastname1}{first1.last1@xxx.edu}
\icmlcorrespondingauthor{Firstname2 Lastname2}{first2.last2@www.uk}

\icmlkeywords{Machine Learning, ICML}

\vskip 0.3in
]

\printAffiliationsAndNotice{}




\fi

\newcommand{\fix}{\marginpar{FIX}}
\newcommand{\new}{\marginpar{NEW}}

\ifdefined\isarxiv
\begin{titlepage}
  \maketitle
  \begin{abstract}
\input{0_abstract}

  \end{abstract}
  \thispagestyle{empty}
\end{titlepage}

{%\hypersetup{linkcolor=black}
%\tableofcontents
}
\newpage

\else

% \maketitle//

\begin{abstract}
\input{0_abstract}
\end{abstract}

\fi

\input{1_intro}
%%% Section 1. Introduction
\input{2_related_work}
\input{3_preli}

\input{4_insight_I}
\input{5_insight_II}
\input{6_insight_III_IV}
\input{7_insight_V}
\input{8_conclusion}

\ifdefined\isarxiv
%\section*{Acknowledgments}
% \bibliographystyle{alpha}
% \bibliography{ref}
\else
%\newpage
% \bibliography{ref}
\bibliographystyle{icml2025}
\bibliography{ref}

\fi

\newpage
\onecolumn
\appendix

\begin{center}
	\textbf{\LARGE Appendix }
\end{center}
{\hypersetup{linkcolor=black}
\tableofcontents
\bigbreak
\bigbreak
\bigbreak
\bigbreak
\bigbreak
}
\newpage

\input{9_app_more_related_work}
\input{10_app_preli}
\input{11_app_concentration}
\input{12_app_defs}
\input{13_app_attention_sparsity}

\input{14_error_analysis}

\ifdefined\isarxiv
\bibliographystyle{alpha}
\bibliography{ref}
\else
\fi

%%%% Cut-line between first 10 pages and appendix

%%% some writing rules

%% Writing rule for creating tags.
%% Tags :
%% Theorem    \ref{thm:bla_bla}
%% Lemma      \ref{lem:bla_bla}
%% Claim      \ref{cla:bla_bla}
%% Corollary  \ref{cor:bla_bla}
%% Fact       \ref{fac:bla_bla}
%% Definition \ref{def:bla_bla}
%% Section    \ref{sec:bla_bla}
%% Subsection \ref{sub:bla_bla}
%% Equation   \ref{eq:bla_bla}

\end{document}

%% file: 0_abstract.tex
% The computational intensity of Large Language Models (LLMs) is a critical bottleneck, primarily due to the $\Theta(n^2)$ complexity of the attention mechanism in transformer architectures. Addressing this, Sparse Attention emerges as a key innovation, aiming to reduce computational load while maintaining model performance. However, as this technique and its variants are widely applied to employ advanced AI systems, we still have a lack of theoretical understanding on when and why it performs competitively to vanilla attention. In this work, 

% This study presents a rigorous theoretical analysis of the sparsity in attention scores within LLMs, particularly under the framework of Gaussian inputs. By establishing a set of foundational assumptions and employing a methodical theoretical approach, we unravel the intrinsic characteristics of attention score sparsity and its implications on computational efficiency. Our main contribution lies in providing a detailed theoretical examination of how sparsity manifests in attention mechanisms, offering insights into the potential trade-offs between computational savings and model effectiveness. This work not only advances our understanding of sparse attention but also provides a scaffold for future research in optimizing the computational frameworks of LLMs, paving the way for more scalable and efficient AI systems.

Sparse Attention is a technique that approximates standard attention computation with sub-quadratic complexity. This is achieved by selectively ignoring smaller entries in the attention matrix during the softmax function computation. Variations of this technique, such as pruning KV cache, sparsity-based fast attention, and Sparse Transformer, have been extensively utilized for efficient Large Language Models (LLMs) deployment. Despite its widespread use, a theoretical understanding of the conditions under which sparse attention performs on par with traditional attention remains elusive. This work aims to {\bf bridge this gap by examining the inherent sparsity of standard attention processes.} Our theoretical framework reveals several brand-new key insights: 
\begin{itemize}
    \item Attention is $n^{C}$-sparse, implying that considering only the largest $\Omega(n^{C})$ entries out of all $n$ entries is sufficient for sparse attention to approximate the exact attention matrix with decreasing loss. Here, $n$ represents the input length and $C \in (0, 1)$ is a constant.
    \item Stable $o(\log(n))$-sparse attention, which approximates attention computation with $\log(n)$ or fewer entries, may not be feasible since the error will persist at a minimum of $O(1)$.
    \item An adaptive strategy ($\alpha \cdot n^C, \alpha\in \R$) for the window size of efficient attention methods rather than a fixed one is guaranteed to perform more accurately and efficiently in a task for inference on flexible context lengths.
\end{itemize}
 
% Here, $n$ represents the input length and $C \in (0, 1)$ is a constant. 2) Stable $o(\log(n))$-sparse attention, which approximates attention computation with $\log(n)$ or fewer entries, may not be feasible. Our results suggest that the $\ell_F$ norm error of approximation to attention scores matrix will persist at a minimum of $O(1)$.

% its additive error will grow by $n^{C_{\rm error}}$ for some $C_{\rm error} \in (0, 1)$.

% for approximating where $n$ is context length and $C \in (0, 1)$; 2) 

% therefore, sparse attention can approximate it just like a knife through butter. Furthermore, we confirm that 1) attention could be approximate $n^C$- sparse for approximating where $n$ is context length and $C \in (0, 1)$; 2) 

% \Chiwun{Here is a todo list
% \begin{itemize}
%     \item Analyze the input distribution of attention and compare it with Gaussian
%     \item Finish the proof of attention sparsity (Done)
%     \item Reorganize all proofs, making them easy to read and to check details (Done)
%     \item Delete Sparse HyperAttention part, discuss the relationships between attention sparsity and fast attention and KV cache in two sections (it will be great if we can contribute more insights)
%     \item Do heavy experiment, analyze sparsity in attention
% \end{itemize}}

% \begin{align*}
%     k_1 = & ~ C_a n_1^C \\
%     k_2 = & ~ C_a n_2^C
% \end{align*}

% \begin{align*}
%     n, \epsilon
%     k
% \end{align*}

%% file: 1_intro.tex
\section{Introduction}\label{sec:intro}

Large Language Models (LLMs) \cite{vsp+17, rns+18, dclt18, rwc+19, bmr+20, cnd+22, zrg+22, cha22} have emerged as a cornerstone of contemporary artificial intelligence, exhibiting remarkable capabilities across a plethora of AI domains. Their prowess is grounded in their ability to comprehend and generate human language with a level of sophistication that is unprecedented. This has catalyzed transformative applications in natural language processing, including machine translation \cite{hwl21}, content creation \cite{cha22, o23}, and beyond, underscoring the profound impact of LLMs on the field of AI.

However, the architectural backbone of these models, particularly those built on the transformer framework \cite{vsp+17}, presents a significant challenge: computational efficiency \cite{tdbm22}. The essence of the transformer architecture, the Attention mechanism, necessitates a computational and memory complexity of $O(n^2)$, where $n$ represents the sequence length. This quadratic dependency limits the scalability of LLMs, especially as we venture into processing longer sequences or expanding model capacities.

In an effort to mitigate this bottleneck, the AI research community has pivoted towards innovative solutions, one of which is sparse attention \cite{cgrs19, cnm19}. Sparse attention mechanisms aim to approximate the results of the full attention computation by selectively focusing on a subset of the input data points. This is typically achieved by omitting certain interactions in the Query and Key multiplications within the attention mechanism, thereby inducing sparsity in the attention matrix. In order to arrive at the goal of preserving the model's performance while alleviating the computational and memory demands, prior works, including pruning KV cache, sparsity-based fast attention, and sparse transformer modeling, demonstrate outstanding efficiencies with $O(n^{1+o(1)})$ (sub-quadratic) complexity and sub-linear memory cache with competitive performance compared with standard attention across various tasks \cite{lwd+23, zsz+24, kmz23, alsy23, lls24, xzc+21, zhdk23,as23_neurips,as24_iclr,as24_arxiv, hjk+23}.

% A noteworthy advancement in this domain is the introduction of the Reformer model \cite{kkl20}, which adeptly reduces the complexity from $O(n^2)$ to $O(n \log n)$ through the adoption of the Locality Sensitive Hashing (LSH) technique. LSH enables the Reformer to efficiently approximate the attention mechanism, significantly curtailing the computational overhead without substantially compromising the model's efficacy.

Despite these advancements, the theoretical underpinnings of sparse attention mechanisms and their implications on model performance and behavior remain an area of active inquiry. In detail, it's not clear when and when not sparse attention can approximate standard attention with a stable error. Also, the sparsity that attention naturally processes, which we call attention sparsity, lacks a strict confirmation of its existence and measurement. Especially, we would like to ask:
\begin{center}
{\it 
    How Sparse Attention Approximates Exact Attention? 
}
\end{center}

% Moreover, sparse attention usually involves extended training

{\bf Our Contributions.}
In this work, we explore the theory of the sparse attention computation problem. Particularly, we first provide a analysis framework that first theoretically confirms the sparsity appears in standard attention. In detailed, our analysis describes the relationships between attention sparsity and input boundary, weights of attention networks and context length. Therefore, we derive several incremental insights based on this framework. In summary, 
\begin{itemize}
    \item We estimate the lower bound on the requirement for $(\epsilon, k)$-sparse softmax vector, proving the {\bf vanilla attention computation is naturally sparse} (see Section~\ref{sec:insight_I_main_result}).
    \item We introduce the concept of {\it attention collapse}, which demonstrates the number of effective entries in attention matrix provably decrease to $1$ or some constant inevitably. (see Section~\ref{sec:attn_collapse})
    \item We give the sufficient lower bound on the window size of stable sparse attention approximating exact attention computation, $\Omega(n^C)$ for constant $C \in (0, 1)$. This further indicates that sparse attention can recover attention outputs from limited $\Omega(n^C)$ entries while achieving a decreasing error. (see Section~\ref{sub:approx_success})
    \item Meanwhile, we also confirm sparse attention approximation from $o(\log(n))$ entries is not enough for stability and extensibility since the lower bound on error will grow with increasing input length. (see Section~\ref{sub:approx_failure}) 
    \item Therefore, we suggest to use adaptive strategy $k = \alpha \cdot n^C, \alpha > 0, C \in (0, 1)$ for the window size of sparse attention rather than the strategy that fixes the window size for any input. The former is proved more efficient within higher approximation performance. (see Section~\ref{sec:dynamic_top_k})
\end{itemize}

%% file: 2_related_work.tex
\section{Related Work}
\paragraph{Sparse and Efficient Transformer.}
In the landscape of attention mechanisms, Vaswani et al. introduced the transformative transformer model, revolutionizing NLP with its comprehensive self-attention mechanism \cite{vsp+17}. Innovations \cite{cgrs19, lsr+19} in sparse attention presented methods to reduce complexity, maintaining essential contextual information while improving computational efficiency. The Reformer \cite{kkl20} utilized Locality Sensitive Hashing to significantly cut down computational demands, enabling the processing of lengthy sequences. Mongoose \cite{clp+20} adapted sparsity patterns dynamically, optimizing computation without losing robustness. \cite{syy21} introduced a learning-to-hash strategy to generate sparse attention patterns, enhancing data-driven efficiency. HyperAttention \cite{hjk+23} refined attention approximation, balancing computational savings with accuracy. Longformer \cite{bpc20} extended transformer capabilities to longer texts through a mix of global and local attention mechanisms. The Performe \cite{cld+20} offered a novel approximation of softmax attention, reducing memory usage for long sequences. Big Bird \cite{zgd+20} combined global, local, and random attention strategies to surmount traditional transformer limitations regarding sequence length. 
\ifdefined\isarxiv
\else
\vspace{-5mm}
\fi

% Other work has simplified attention computation, like \cite{bsz23,swy23,gsy23_coin,dms23,gsx23_incontext,syz23,zhdk23,as23}. Finally, \cite{gswy23} reformulated attention using tensors and SVMs. Attention has been widely applied, like in graph neural networks by \cite{vcc+17}, in image captioning by \cite{xbk+15}, and in Transformers for NLP by \cite{vsp+17}.
\paragraph{Theoretical Approaches to Understanding LLMs.}
There have been notable advancements in the field of regression models, particularly with the exploration of diverse activation functions, aiding in the comprehension and optimization of these models. The study of over-parameterized neural networks, focusing on exponential and hyperbolic activation functions, has shed light on their convergence traits and computational benefits \cite{bsz23,swy23,gsy23_coin,dms23,gsx23_incontext,syz23,zhdk23,as23_neurips,as24_iclr,as24_arxiv, gswy23, dsxy23, lswy23, csy24}. Enhancements in this area include the addition of regularization components and the innovation of algorithms like the convergent approximation Newton method to improve performance \cite{lsz23}. Additionally, employing tensor methods to simplify regression models has facilitated in-depth analyses concerning Lipschitz constants and time complexity \cite{gsx23_incontext, dlms23}. Concurrently, there's a burgeoning interest in optimization algorithms specifically crafted for LLMs, with block gradient estimators being utilized for vast optimization challenges, significantly reducing computational load \cite{clmy21}. Novel methods such as Direct Preference Optimization are revolutionizing the tuning of LLMs by using human preference data, circumventing the need for traditional reward models \cite{rsm+23}. Progress in second-order optimizers is also notable, offering more leniency in convergence proofs by relaxing the usual Lipschitz Hessian assumptions \cite{llh+23}. Moreover, a series of studies focus on the intricacies of fine-tuning \cite{mgn+23, mwy+23, psza23}. 
These theoretical developments collectively push the boundaries of our understanding and optimization of LLMs, introducing new solutions to tackle challenges like the non-strict Hessian Lipschitz conditions.

%% file: 3_preli.tex
\section{Preliminary}

In this section, we formalize the problem we strive to address, along with the definitions, and assumptions that underpin our work. 
% In Section~\ref{sub:layer_normalization}, we define Layer Normalization \cite{bkh16} as a mild assumption in this work. 
% We formally give the definition of attention computation in Section~\ref{sub:def_attn}. 
Especially, Section~\ref{sub:sparsity} explains the main theory of the problem we aim to study. In Section~\ref{sub:sparse_attn}, we give a unified definition of the sparsity-based attention method and how it approximates exact attention computation, and we call it sparse attention.

{\bf Assumption.} In this work, we consider one-layer self-attention computation both in standard form and sparsity-based approximate form. To begin with, we give the assumption of the input matrix of attention computation, denoted as $X \in \R^{n \times d}$ where $n$ is the context length and $d$ stands the dimension, as follows (refer to Definition~\ref{def:X} for the formal and detailed version of assumption):
\begin{itemize}
    \item {\bf Independent Entries.} For any two entries $X_{i_1, j_1}$ and $X_{i_2, j_2}$ in matrix $X$, $\forall i_1, i_2 \in [n]$ and $j_1, j_2 \in [d]$, they are independent.
    \item {\bf Bounded Entries.} For failure probability $\delta \in (0, 0.1)$. With a probability $1 - \delta$, the entry $X_{i, j}$ in matrix $X$, $\forall i \in [n]$ and $j\in [d]$, we have $| X_{i, j}  | \leq B$ for some positive constant $B > 0$.
\end{itemize}

{\bf Attention Computation.}
Hence, we are about to introduce the standard attention computation, which occupies the main time and space complexity $O(n^2)$ in LLMs inference. First, we denote the weights of query, key and value projection as $W_Q, W_K, W_V \in \R^{d \times d}$. Thus, we let query, key and value state matrices be computed by $Q := XW_Q, K:= XW_K, V:= XW_V \in \R^{n \times d}$. We state the following definition:
\begin{definition}[Attention computation]\label{def:attn:informal}
    Given Query, Key and Value states matrices $Q, K, V \in \R^{n \times d}$. We then define $A := \exp(Q K^\top / \sqrt{d})$, $D := \diag(A{\bf 1}_n)\in \R^{n \times n}$.
    The attention computation ${\sf Attn}(Q, K, V) \in \R^{n \times d}$ is given by:
    \begin{align*}
        {\sf Attn}(Q, K, V) := D^{-1} A V \in \R^{n \times d}
    \end{align*}
    Specially, we denote $D^{-1} = \diag(1/(A {\bf 1}_n))\in \R^{n \times n}$.
\end{definition}

\subsection{Attention Sparsity}\label{sub:sparsity}
In Definition~\ref{def:attn:informal}, $ D^{-1}A \in \mathbb{R}^{n \times n} $ represents the attention matrix, indicating how much the model focuses on each vector. In much of the sparse attention literature, $ D^{-1}A $ is assumed to be sufficiently sparse, allowing sparsity-based efficient attention methods to disregard some zero entries in order to achieve a balance between accuracy and efficiency. In this paper, we introduce a threshold, denoted as $ \epsilon $, and define {\bf attention sparsity} as {\bf the number of entries in each row of $D^{-1}A \in \mathbb{R}^n$ that are smaller than  $\epsilon$}. Specifically, for a softmax vector $ u \in \mathbb{R}^n $, if there are at least $ n-k $ entries in $ u $ that are not greater than $ \epsilon $ for all integers $ k \in [n] $, we say that $ u $ is $(\epsilon, k)$-sparse. Since $ \epsilon $ is intended to be a very small value, we will simply refer to $ u $ as being $ k $-sparse. The formal definition is provided below:

\begin{definition}[$(\epsilon, k)$-sparsity]\label{def:S:informal}
    For a vector $u \in \R^{n}$ and error $\epsilon > 0$, we define sparse set $\mathcal{S}_\epsilon(u)$ as:
    $
        \mathcal{S}_{\epsilon}(u) := \{ i \in [n] ~|~ |u_i| \le \epsilon \}
    $.
    Hence, we say $u$ is at least $(\epsilon, k)$-sparse when it holds that $| \mathcal{S}_{\epsilon}(u) | \ge n-k.$
\end{definition}
{\bf Problem Definition I: Estimating $\epsilon$.} In practical implementations, establishing a clear relationship between $ \epsilon $ and the sparsity $ k $ proves to be challenging. Therefore, we first analyze how to estimate a boundary for $ \epsilon $ based on a given sparsity integer $ k \in [n] $. Naively, given $k$, we would like to find a guaranteed value for $\epsilon$ that satisfies $| \mathcal{S}_{\epsilon}(u) | \ge n-k$. Addression this problem will enable us to assess the loss associated with approximating standard attention using sparse attention, ultimately guiding us in finding the optimal trade-off between $ \epsilon $ (where lower values yield greater accuracy) and $ k $ (where lower values lead to higher efficiency).

\subsection{Sparse Attention and Approximation}\label{sub:sparse_attn}
Here we state an ideal mathematical definition for the sparse attention in this paper. Initially, we define a set, ${\cal T}_{k}(u)$, to filter out the greatest $k$ entries in a vector $u \in \R^n$. The integer $k \in [1, n]$ is also called window size in some sparse attention works.
\begin{definition}\label{def:topk:informal}
    For a vector $u \in \R^{n}$, given a sparsity integer $k$, we denote a top-$k$ set ${\cal T}_{k}(u) := \{ i \in [n] ~|~ {\cal S}_{u_i}(u) \ge n - k \}$, then we define vector
    \begin{align*}
        {\sf topk}(u) := [ u_1 \cdot {\bf 1}_{1 \in {\cal T}_{k}(u)}, \cdots, u_n \cdot {\bf 1}_{n \in {\cal T}_{k}(u)} ]^\top \in \R^n.
    \end{align*}
\end{definition}

Note that ${\bf 1}_{i \in {\cal T}_{k}(u)}$ is an indicator where when $i \in {\cal T}_{k}(u)$, it equals $1$, otherwise, $0$. We utilize ${\sf topk}(u)$ to compute a sparsity-based approximating version of $A = \exp(QK^\top)$ in Definition~\ref{def:attn:informal}, we denote it $A_{\rm spar}$. Accordingly, we provide a universal version for all sparsity-based attention as follows:
\begin{definition}[Sparse attention]\label{def:sparse_attn:informal}
    Given Query, Key and Value state matrices $Q, K, V \in \R^{n \times d}$. We then define $A := \exp(Q K^\top / \sqrt{d}) \in \R^{n \times n}$. Especially, we define $A_{\rm spar} := \begin{bmatrix}
        {\sf topk}(A_{1, *}), \cdots, {\sf topk}(A_{n, *})
    \end{bmatrix}^\top$,
    $D_{\rm spar} := \diag(A_{\rm spar} {\bf 1}_n)\in \R^{n \times n}$.
    The sparse attention computation ${\sf SparseAttn}(Q, K, V) \in \R^{n \times d}$ is given by:
    \begin{align*}
        {\sf SparseAttn}(Q, K, V) := D_{\rm spar}^{-1} A_{\rm spar} V \in \R^{n \times d}
    \end{align*}
    Specially, we denote $D_{\rm spar}^{-1} = \diag(1/(A_{\rm spar} {\bf 1}_n))\in \R^{n \times n}$.
\end{definition}
It should be noted that directly accessing top k entries in the attention matrix without any extra computational cost is overly ideal for efficient LLMs in real-world cases. Prior works usually utilize some additional approximate algorithm to meet this condition, e.g. Locality-Sensitive Hashing (LSH) for retrieving larger query-key pairs, but this also brings more approximating errors. We only focus on the part of approximating attention computation in this study and leave the part of pre-approximating top-k entries in $D^{-1} A$ as a future direction.

{\bf Problem Definition II: Sparse Attention Approximation.} The variations of sparse attention, including pruning KV Cache \cite{llc+21, xtc+23} and sparsity-based attention \cite{kkl20, zhdk23, hjk+23}, focus on solving the approximation of the attention matrix, where we call it sparse attention approximation. In particular, we emphasize the importance of {\it stable sparse attention approximation}, which directly affects the extensibility of sparse attention under long context scenes. We denote $f: \mathbb{N}^+ \rightarrow \mathbb{N}^+$ as the strategy to choose a suitable window size due to different input lengths. Hence, we give:
\begin{definition}[{\it Stable sparse attention approximation} ${\sf SSAA}(f)$]\label{def:ssaa}
    For some strategy $f:\mathbb{N}^+ \rightarrow \mathbb{N}^+$ to choose the sparsity $k = f(n)$ in sparse attention (Definition~\ref{def:sparse_attn:informal}), the problem of stable sparse attention approximation ${\sf SSAA}(f)$ is to solve:
    \begin{align*}
        L(f, n) = \| D_{\rm spar}^{-1} A_{\rm spar} - D^{-1} A \|_p,
    \end{align*}
    where $\| \cdot \|_p$ denotes some norm. We say this sparse attention approximation is {\bf stable} iff:
    \begin{itemize}
        \item $L(f, n)$ is monotonically decreasing with growing $n$.
        \item $\lim_{n \rightarrow +\infty} L(f, n) = 0$.
    \end{itemize}
\end{definition}

%% file: 4_insight_I.tex
\section{Theoretical Insight I: Attention is Provably Naturally Sparse}\label{sec:insight_I_main_result}

In this section, we expound on our principal theorem that delineates the sparsity attention, accompanied by a lower limit on $\epsilon$. Specifically, the sparsity of attention could be influenced by numerous factors intrinsic to attention computation, such as the length of the context, the weightage assigned to projections of attention, as well as the vector similarity between Query and Key states. Our objective is to identify any inherent principles that shed light on the attention sparsity, consequently aiding the development of efficient attention algorithms.

Initially, we posit that under our assumptions in the previous section, each element in the vector $QK^\top$ is bounded with a high probability, as detailed in Section~\ref{sub:concentrations}. Subsequently, we reinterpret our main problem as a probabilistic problem and give the estimation for $\epsilon$ to establish the $(\epsilon, k)$-sparsity for a given sparsity integer $k \in [n]$ of attention in Section~\ref{sub:main_result}.

\subsection{Concentrations}\label{sub:concentrations}

Before we give our concentration results in attention, firstly, we introduce a key operator in our work, where we conclude that most of the factors influence attention sparsity into one variable. In Definition~\ref{def:R:informal}, $R \ge 0$ is defined by two main factors that may bring negative effects to attention sparsity, where we show as follows:
\begin{definition}\label{def:R:informal}
    Denote $W := W_Q W_K^\top /\sqrt{d} \in \R^{d \times d}$. We define 
    $
        R := B^2 \cdot \| W\|_F
    $.
\end{definition}

Hence, the theoretical concentration results are given below.
\begin{lemma}\label{lem:concentrations:informal}
    $\delta \in (0, 0.1)$. Let $R \ge 0$ be defined as Definition~\ref{def:R:informal}. $\forall i_1, i_2 \in [n]$. Then with a probability at least $1 - \delta$, we have
    \begin{itemize}
        \item Part 1. $|(QK^\top)_{i_1, i_2}| \leq O(R) \cdot \sqrt{\log(d/\delta)}$.
        \item Part 2. $\exp(- O(R) \cdot \sqrt{\log(d/\delta)}) \leq A_{i_1, i_2} \leq \exp( O(R) \cdot \sqrt{\log(d/\delta)})$.
        \item Part 3. $D_{i_1, i_1}^{-1} \leq \exp( O(R) \cdot \sqrt{\log(nd/\delta)}) / n$.
    \end{itemize}
\end{lemma}

\begin{proof}[Proof sketch of Lemma~\ref{lem:concentrations:informal}]
    Proof of Part 1 follows from the property of inputs $X$ is bounded, then we apply the Hoeffding inequality (Lemma~\ref{lem:hoeffding}) to have the results. Proofs of Part 2 and Part 3 are just a simple extension of Part 1 in attention computation. Please refer to Lemma~\ref{lem:QK_concen} for the formal proof of Part 1 and refer to Lemma~\ref{lem:bound_D} for the formal proof of Part 2.
\end{proof}

{\bf Remark.} The formal results of Lemma~\ref{lem:concentrations:informal} in the appendix have slight differences with the informal forms, in which we omit the additional term of each upper bound since such terms are trivially some constants. Fact~\ref{fac:softmax_bias} shows any constant bias term added before the softmax function will not change the output. We thus simplify the equations for tighter boundaries and more convenient notation. (See Remark~\ref{rem:addtional_term})

\subsection{Attention Sparsity with Bound on Error}\label{sub:main_result}

We state our result in Theorem~\ref{thm:main_result:informal}.
\begin{theorem}\label{thm:main_result:informal}
    Let $R \ge 0$ be defined as Definition~\ref{def:R:informal}. Given sparsity integer $k \le n$. Denote $T := \exp(\sqrt{\log(n(n-k)d/\delta)})$. Let $\mathcal{S}_\epsilon$ be defined as Definition~\ref{def:S:informal}. $\delta \in (0, 0.1)$. If we choose $\epsilon \ge \frac{T^{O(R)}}{n}$,
    then with a probability at least $1 - \delta$, for all $i \in [n]$, we have
    $
        \Big| \mathcal{S}_\epsilon(D^{-1}_{i, i} A_{i, *}) \Big| \ge n - k
    $.
\end{theorem}

\begin{proof}[Proof sketch of Theorem~\ref{thm:main_result:informal}]
    Follows from Definition~\ref{def:S:informal}, there is less than $k$ entries in $D^{-1}_{i, i} A_{i, *} \in \R^n$ that are greater than $\epsilon$. We can easily get $\epsilon \ge \frac{T^{O(R)}}{n}$ by combining Part 2 and Part 3 of Lemma~\ref{lem:concentrations:informal}. The complete proof is provided in Appendix~\ref{sec:main_result} and Theorem~\ref{thm:main_result}.
\end{proof}

Theorem~\ref{thm:main_result:informal} gives the relationship between attention sparsity $k$ and the error of approximating attention computation by considering only top-$k$ entries in attention. To our best knowledge, it is the {\bf first theoretical analysis} that confirms the attention sparsity. It provides a vital framework for our further analysis of sparse attention approximation on exact attention computation. We do not conduct any empirical evaluation for this theorem since this is a trivial finding in the field of efficient attention computations.

%% file: 5_insight_II.tex
\section{Theoretical Insight II: Attention Collapse}\label{sec:attn_collapse}

This section introduces the concept of {\it attention collapse}, we formally define it as a phenomenon in attention computation that reveals {\bf whatever the values of $X$, $B$ and entries of $W_Q$ and $W_K$ are (also could be considered as $X$ and $R$), the number of effective entries, whose values are bigger than $\epsilon$, in each row of attention matrix, will inevitably decrease to 1 or some constant due to a considerable large input length $n$}. This result only requires a simple derivation from Theorem~\ref{thm:main_result}, as we show:
\begin{theorem}\label{thm:attn_collapse:informal}
    Consider a fixed $\epsilon$ with a very small value, $\delta \in (0, 0.1)$. Then with a probability at least $1- \delta$, there is:
    \begin{itemize}
        \item Part 1. If $R = o(\sqrt{\log(n)})$, then we have $\lim_{n \rightarrow +\infty} | \mathcal{S}_{\epsilon}(u) | \ge n - 1$.
        \item Part 2. If $R = O(\sqrt{\log(n)})$, then we have $\lim_{n \rightarrow +\infty} | \mathcal{S}_{\epsilon}(u) | \ge O(1)$.
        % \item Part 3. If $R = \Omega(\sqrt{\log(n)})$, then we have $\lim_{n \rightarrow +\infty} | \mathcal{S}_{\epsilon}(u) | = 0$.
    \end{itemize}
    % \begin{align*}
    % $
    %     \lim_{n \rightarrow +\infty} | \mathcal{S}_{\epsilon}(u) | = n - 1.
    % $
    % \end{align*}
\end{theorem}

\begin{proof}[Proof sketch of Theorem~\ref{thm:attn_collapse:informal}]
    Consider a fixed $\epsilon$ with a very small value, $| \mathcal{S}_{\epsilon}(u) |$ could be obtained by $n - k$, where we have:
    \begin{align}\label{eq:S}
        | \mathcal{S}_{\epsilon}(u) | \ge \exp\Big( O(\frac{\log^2(\epsilon \cdot n)}{R^2}) \Big) \cdot \frac{\delta}{nd}.
    \end{align}
    {\bf Proof sketch of Part 1:}
    Knowing $\epsilon, \delta, R, d$ are fixed, this equation will grow hyper-linear with an increasing $n$, its maximal is at least $n-1$ because of some properties of the softmax function. 
    {\bf Proof sketch of Part 2:}
    When $R = O(\sqrt{\log(n)})$, the lower bound on $| \mathcal{S}_{\epsilon}(u) |$ is some constant that doesn't relate to $n$. 
    % When $R = \Omega(\sqrt{\log(n)})$, the lower bound on $| \mathcal{S}_{\epsilon}(u) |$ will grow with an increasing $n$.
    Refer to Theorem~\ref{thm:attn_collapse} for the detailed proof.
\end{proof}

{\bf Discussion.} Following Theorem~\ref{thm:attn_collapse}, we find that there is a crucial parameter for the attention sparsity, $R = B^2\| W\|_F$. When $R = o(\sqrt{\log(n)})$ or $R = O(\sqrt{\log(n)})$, which satisfies the bound for most of the situations in current transformer-based models, {\it attention collapse} inevitably happens when inputting super-long matrix into attention network. Due to its natural limitation, attention can only focus on a few entries even if we are still inputting more pieces of information, we believe this provides some explanations for several past discoveries, for instance, attention sink \cite{xtc+23}, low-rank preserve of attention \cite{cam+24}.

Moreover, this theorem can be connected to the fine-grained complexity of attention computation - the bound is astonishingly similar! \cite{as23_neurips} confirm that any fast attention algorithm approximates standard attention within $n^{1+o(1)}$ executions and $1/\poly(n)$ error requires query, key and value state matrices are bounded, mathematically $\|Q\|_\infty \leq o(\sqrt{\log(n)})$, $\|K\|_\infty \leq o(\sqrt{\log(n)})$, $\|V\|_\infty \leq o(\sqrt{\log(n)})$. Since the $\ell_F$ norm can be transformed into $\ell_\infty$ norm, our results further validate the effectiveness of efficient attention theoretically from the perspective from sparsity.
% $\sqrt{\log(n)}$ for the input may be the most important 

{\bf Experimental Evaluation.} Figure~\ref{fig:attn_collapse} indicates the effectiveness of our theory: For a fixed $R$, we denote the number of entries that are small than or equal $\varepsilon$ in attention matrix as $|{\cal S}_\varepsilon|$. We sufficiently sample input $X \in \R^{n \times d}$ from Gaussian distribution with various input length $n \in \{ 2^1, 2^2, \cdots, 2^{24} \}$. For different $\varepsilon \in \{{\rm 1e-4}, 5e-5, 2e-5, 1e-5, \cdots, 5e-7, 2e-7, 1e-7\}$, $|{\cal S}_\varepsilon| / n$ inevitably decreases to 0. 
% Please refer to Appendix~\ref{??} for the details of this experiment.

\ifdefined\isarxiv
\begin{figure}
    \centering
    \includegraphics[width=0.5\textwidth]{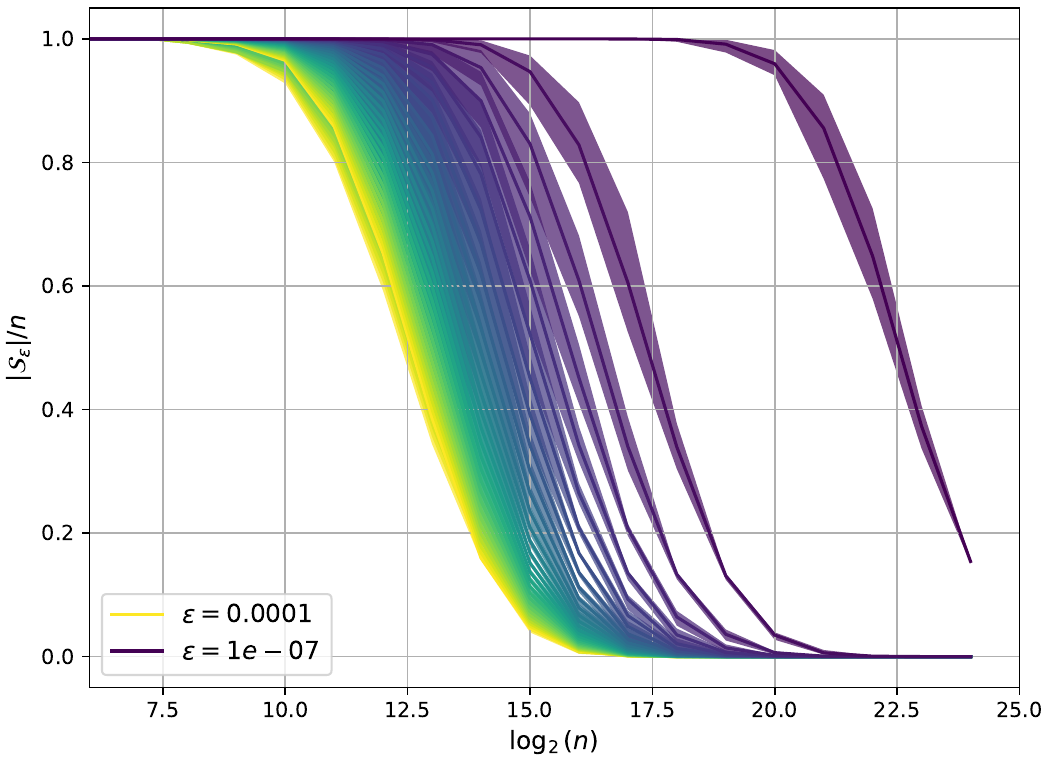}
    \caption{Count of ineffective entries divided by $n$ in attention computation, denoted as ${\cal S}_{\varepsilon}/n$, decreases with an increasing $n$, where $\varepsilon$ is choosing from $\{{\rm 1e-4}, 5e-5, 2e-5, 1e-5, \cdots, 5e-7, 2e-7, 1e-7\}$. }
    \label{fig:attn_collapse}
\end{figure}
\else
\begin{figure}
    \centering
    \includegraphics[width=\linewidth]{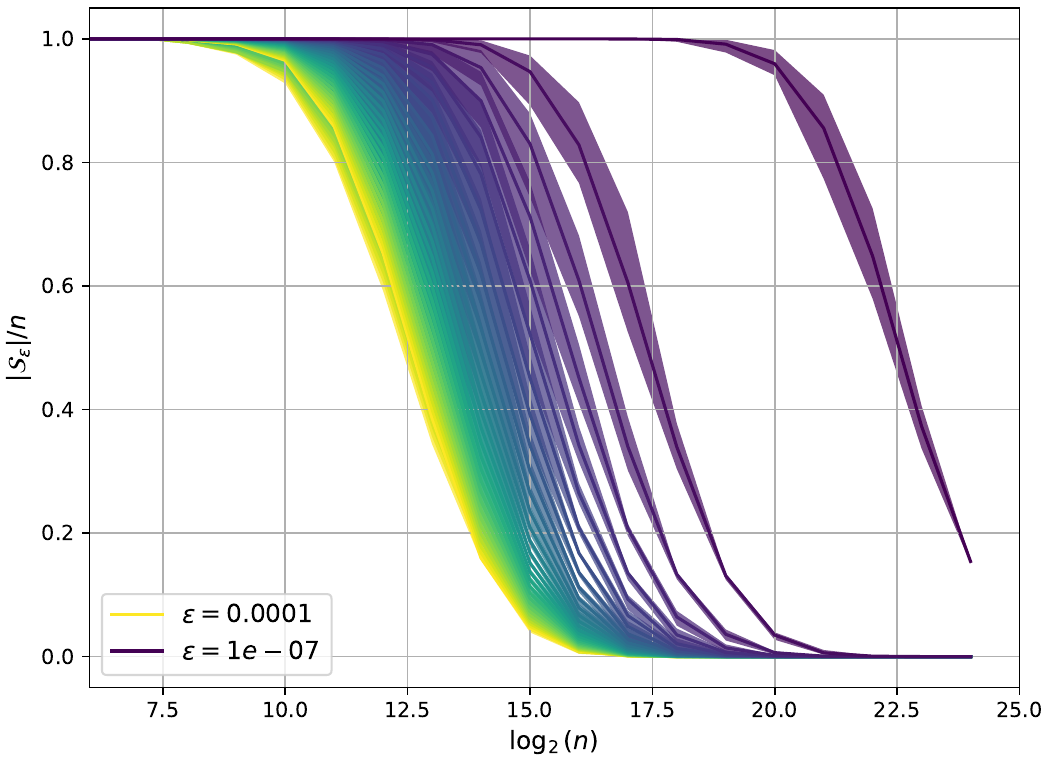}
    \caption{Count of ineffective entries divided by $n$ in attention computation, denoted as ${\cal S}_{\varepsilon}/n$, decreases with an increasing $n$, where $\varepsilon$ is choosing from $\{{\rm 1e-4}, 5e-5, 2e-5, 1e-5, \cdots, 5e-7, 2e-7, 1e-7\}$. }
    \label{fig:attn_collapse}
    \vspace{-7mm}
\end{figure}
\fi

%% file: 6_insight_III_IV.tex
\begin{figure*}
    \centering
    \includegraphics[width=\linewidth]{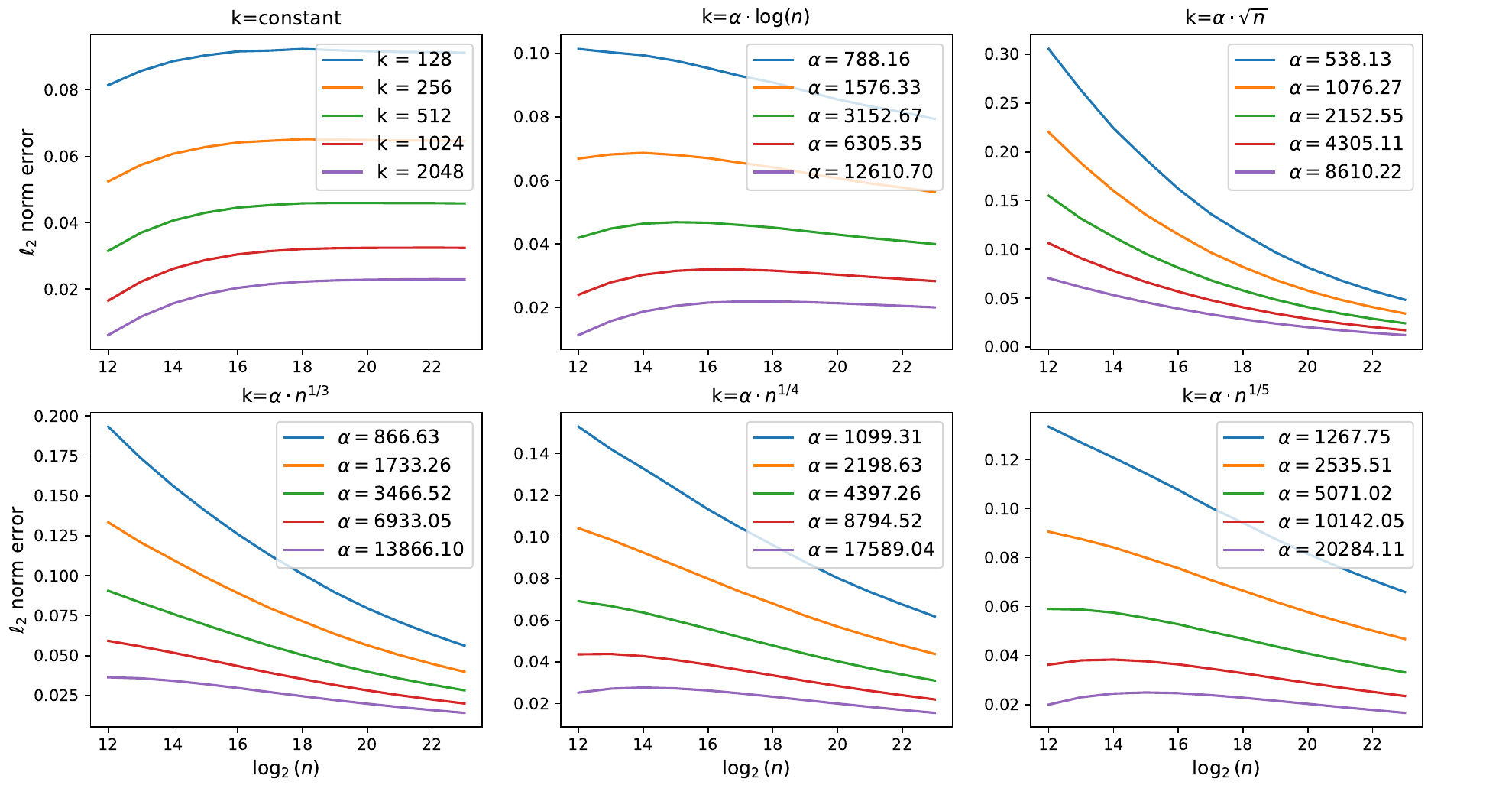}
    \caption{Simulation of sparse attention approximation with different window size settings $k$. The $\ell_2$ norm error means the average approximating error on the attention matrix ($\frac{1}{n} \| D_{\rm spar}^{-1} A_{\rm spar} - D^{-1} A\|_2$). Note that the compute complexities of the lines with the same colors is equal.}
    \label{fig:ssaa_simulation}
\end{figure*}

\section{Theoretical Insight III and IV: On Stable Sparse Attention Approximation}\label{sec:III_&_IV}

The goal of this section is to explain the success of sparse attention approximation as we defined in Definition~\ref{def:ssaa}, meanwhile, our analysis indicates an incremental sufficient requirement for {\it stable sparse attention approximation}. In detail, we provide the minimum low bound for the window size $k$ in sparse attention for ${\sf SSAA}$ in Section~\ref{sub:approx_success}. On contrary, we give the lower bound on approximating the error of some common settings in prior works of sparse attention in Section~\ref{sub:approx_failure}, which we aim to theoretical support that an invariable $k$ or even $k = o(\log(n))$ is inappropriate for more efficient {\it stable sparse attention approximation}. Finally, in Section~\ref{sub:approx_eval} we conduct experiments to evaluate the effectiveness of our conclusions.

\subsection{Theoretical Insight III: \texorpdfstring{$\Omega(n^C)$}{} Entries Suffice for Stable Sparse Attention Approximation}\label{sub:approx_success}

Based on the attention sparsity framework in Theorem~\ref{thm:main_result:informal}, we showcase the sufficient bound for ${\sf SSAA}$, which is:
\begin{theorem}\label{thm:approx_success:informal}
    $\delta \in (0, 0.1)$. For a constant $C \in (0, 1)$, we then denote $f(n) := \Omega(n^C)$, therefore, with a probability at least $1 - \delta$, window size strategy $k = f(n)$ is sufficient to solve ${\sf SSAA}(f)$ in Definition~\ref{def:ssaa}.
\end{theorem}
% \vspace{-5mm}

\begin{proof}[Proof sketch of Theorem~\ref{thm:approx_success:informal}]
    Recall that $L(f, n) = \| D_{\rm spar}^{-1} A_{\rm spar} - D^{-1} A \|_p$ for $p = \infty$, we have $D_{\rm spar}^{-1} A_{\rm spar} - D^{-1} A \|_\infty \leq \| D_{\rm spar}^{-1} A_{\rm spar} - D^{-1} A_{\rm spar} \|_\infty + \| D^{-1} A_{\rm spar} - D^{-1} A \|_\infty$. Thus, these two terms are bounded following Part 1 and Part 4 of Lemma~\ref{lem:upper_bound_error}, we can obtain that $L(f, n) \leq 1 / n^C_{\rm error}$ for a constant $C_{\rm error} < C$ when $k = f(n) = n^C$ for a constant $C \in (0, 1)$.
\end{proof}

\subsection{Theoretical Insight IV: Sparse Attention Fail to Approximate Exact Attention from \texorpdfstring{$o(\log(n))$}{} Entries}\label{sub:approx_failure}

Since we provide a sufficient bound for the window size in sparse attention in Theorem~\ref{thm:approx_success:informal}, we also give the insufficient bound which leads the failure of {\it stable sparse attention approximation} as follows:
\begin{theorem}\label{thm:approx_failure:informal}
    $\delta \in (0, 0.1)$. For a constant $C \in (0, 1)$, we then denote $f(n) := o(\log(n))$, therefore, with a probability at least $1 - \delta$, window size strategy $k = f(n)$ cannot solve ${\sf SSAA}(f)$ in Definition~\ref{def:ssaa}.
\end{theorem}

\begin{proof}[Proof sketch of Theorem~\ref{thm:approx_failure:informal}]
    This proof follows from choosing $p$ as Frobenius norm and $k = o(\log(n))$, the error function then will be at least monotonically increasing with $n$. Please refer to Theorem~\ref{thm:approx_failure} for the formal version and corresponding detailed proofs.
\end{proof}

{\bf Discussion.} Theorem~\ref{thm:approx_success:informal} is an incremental result to theoretically support the existence of attention sparsity. Different from Theorem~\ref{thm:main_result:informal}, Theorem~\ref{thm:approx_success:informal} gives the sufficient bound for the number of entries that {\it stable sparse attention approximation} requires. Therefore, we ensure that attention computation is at least $(\epsilon, \Omega(n^C))$-sparse with a very small $\epsilon > 0$. This provides a new perspective on why sparse attention can sometimes be competitive with standard attention. Theorem~\ref{thm:approx_failure:informal} reveals another side of the coin - $o(\log(n))$ window size is not enough for {\it stable sparse attention approximation}, which demonstrates a theoretical perspective for current efficient KV cache methods might still lack scalability towards super-long context length.

\subsection{Empirical Evaluation}\label{sub:approx_eval}

{\bf Setup.} This section provides an empirical evaluation of our theoretical results in Section~\ref{sub:approx_success} and Section~\ref{sub:approx_failure}. Here we aim to give some simple examples to confirm our theory. In detail, we simulate the error dynamics of sparse attention approximation within different window size strategies. For instance, we compare the approximating error on attention matrix when $k = C_1$, $k = \alpha_1 \cdot \log(n)$ and $k = \alpha_2 \cdot n^{C_2}$, where $C_1, C_2, \alpha_1$ and $\alpha_2$ are all constants. Specially, we consider a one-layer attention network with $W_Q = W_K = \diag( {\bf 1}_d)$. We choose input length $n$ from $\{2^{12}, 2^{13}, \cdots, 2^{23}\}$. For input $X \in \R^{n \times d}$, we sample each entry from Gaussian distribution $\mathcal{N}(0, 1)$. We use constant window size as the baseline of this experiment, where $k$ will be chosen from $\{128, 256, 512, 1024, 2048\}$. For dynamic window size strategy $k= \lfloor \alpha \cdot \log(n) \rfloor$ and $k = \lfloor \alpha \cdot n^C \rfloor$, we adjust $\alpha$ to enforce the computational cost is aligned to the cost when applying fixed window size. For example, the computational cost of $k = 128$ is $O(\sum_{i=1}^{12} 128 n_i)$, we compute $\alpha_1 := \sum_{i=1}^{12} 128 n_i / \sum_{i=1}^{12} \log(n_i) n_i$ and $\alpha_2 := \sum_{i=1}^{12} 128 n_i / \sum_{i=1}^{12} \sqrt{n_i} n_i$ to satisfy that the compute complexity of $k= \lfloor \alpha_1 \cdot \log(n) \rfloor$ and $k= \lfloor \alpha_2 \cdot \sqrt{n} \rfloor$ equal to $O(\sum_{i=1}^{12} 128 n_i)$. We repeat 1000 times for each setting and take the mean value for the stable and fair result.

{\bf Results.} As Figure~\ref{fig:ssaa_simulation} shows, the error of sparse attention approximation with $k=C$ and $k = O(\log(n))$ window size strategy grows or maintains under the trend of an increasing context length. Meanwhile, window size strategies with $n^C$ functions perform descents on the error with growing input length $n$. Moreover, since the total compute costs are aligned the same for different window size strategies, we validate their mean values of approximating errors and record them in Table~\ref{tab:attn_approx_simulation}. As shown, the constant window size strategy performs well, and the $k = \alpha n^{1/1000}$ strategy performs with minimal error, which indicates the dynamic strategy for the window size might be a better choice for the design of advanced sparse attention approaches. We further explore this problem in Section~\ref{sec:dynamic_top_k}.

\begin{table}[t]
    \centering
   % \resizebox{0.5\linewidth}{!}{
   \begin{tabular}{@{}ccc@{}}
   \toprule
   \multirow{1}{*}{Window Size Strategy} & \multirow{1}{*}{Average $\frac{1}{n}\| D_{\rm spar}^{-1} A_{\rm spar} - D^{-1} A\|_2$} \\
   \midrule
   constant & $0.04977$ \\
   $\log(n)$ & $0.05080$ \\
   $\sqrt{n}$ & $0.08504$ \\
   $n^{1/100}$ & $0.04974$ \\
   $n^{1/1000}$ & ${\bf 0.04972}$ \\
   \bottomrule
   \end{tabular}
  % }
   % \label{tab:attn_approx_simulation}
   \caption{Mean error of different window size strategies on our synthetic data and one-layer attention computation. In particular, we bold the minimal error as the best performance for {\it stable sparse attention approximation} in these strategies. }
   \label{tab:attn_approx_simulation}
\end{table}

%% file: 7_insight_V.tex
\begin{table*}[t]
\vspace{-1mm}
\centering
\resizebox{\textwidth}{!}{
\scriptsize
\begin{tabular}{c@{}c@{}c@{}c@{} c@{}c@{}c@{} c@{}c@{}c@{} c@{}c@{}c@{} c@{}c@{} c@{}c@{} c}
% \toprule
\multirow{2}{*}{\raisebox{-4ex}{\textbf{Methods}}}  % Moves the text down
& \multicolumn{3}{c}{\textbf{Single-Document QA}} 
& \multicolumn{3}{c}{\textbf{Multi-Document QA}} 
& \multicolumn{3}{c}{\textbf{Summarization}} 
& \multicolumn{3}{c}{\textbf{Few-shot Learning}} 
& \multicolumn{2}{c}{\textbf{Synthetic}} 
& \multicolumn{2}{c}{\textbf{Code}} 
& \multirow{2}{*}{\raisebox{-4ex}{\textbf{Avg.}}} 
% & \multirow{2}{*}{\raisebox{-4ex}{\textbf{ Time}}} 
% & \multirow{2}{*}{\raisebox{-6ex}{\textbf{\shortstack{Time \\ (s / sample)}}}}
\\  % Moves the text down

\cmidrule(lr){2-4} \cmidrule(lr){5-7} \cmidrule(lr){8-10} \cmidrule(lr){11-13} \cmidrule(lr){14-15} \cmidrule(lr){16-17}
\setlength{\tabcolsep}{1pt} % 默认值是6pt，减小这个值来减少列间距
& \makebox[1cm]{\raisebox{0.5ex}{\rotatebox{30}{\textbf{NtrvQA}}}} 
& \makebox[1cm]{\raisebox{0.7ex}{\rotatebox{30}{\textbf{Qasper}}}} 
& \makebox[1cm]{\raisebox{0.8ex}{\rotatebox{30}{\textbf{MF-en}}}} 
& \makebox[1cm]{\raisebox{0.4ex}{\rotatebox{30}{\textbf{HotpotQA}}}} 
& \makebox[1cm]{\raisebox{0.3ex}{\rotatebox{30}{\textbf{2WikiMQA}}}} 
& \makebox[1cm]{\raisebox{0.7ex}{\rotatebox{30}{\textbf{Musique}}}} 
& \makebox[1cm]{\raisebox{0.5ex}{\rotatebox{30}{\textbf{GovReport}}}} 
& \makebox[1cm]{\raisebox{0.8ex}{\rotatebox{30}{\textbf{QMSum}}}} 
& \makebox[1cm]{\raisebox{0.6ex}{\rotatebox{30}{\textbf{MultiNews}}}} 
& \makebox[1cm]{\raisebox{0.8ex}{\rotatebox{30}{\textbf{TREC}}}} 
& \makebox[1cm]{\raisebox{0.6ex}{\rotatebox{30}{\textbf{TriviaQA}}}} 
& \makebox[1cm]{\raisebox{0.6ex}{\rotatebox{30}{\textbf{SAMSum}}}} 
& \makebox[1cm]{\raisebox{0.6ex}{\rotatebox{30}{\textbf{PCount}}}}  
& \makebox[1cm]{\raisebox{1.4ex}{\rotatebox{30}{\textbf{PRe}}}}    
& \makebox[1cm]{\raisebox{1.6ex}{\rotatebox{30}{\textbf{Lcc}}}} 
& \makebox[1cm]{\raisebox{1.4ex}{\rotatebox{30}{\textbf{RB-P}}}} \\

% Llama-2-7B-chat-hf
\midrule
\multicolumn{18}{c}{\textbf{LlaMa-3-8B-Instruct  }} \\
% \arrayrulecolor[gray]{0.8}
\midrule
% \arrayrulecolor{black}
% \rowcolor{lightgray} 
FullKV & 25.70 & 29.75 & 41.12 & 45.55 & 35.87 & 22.35 & 25.63	& 23.03&	26.21&	73.00	&90.56	&41.88	&4.67&	69.25&	58.05	&50.77	& 41.46 \\
\midrule
% \multicolumn{18}{c}{\textbf{Llama-2-7B-chat-hf, KV Size = 384 , Compressibility is 9.38\%  (Except CHAI method)}} \\
% \arrayrulecolor[gray]{0.8}
% \midrule
% \arrayrulecolor{black}
StreamingLLM (k=2048) & 21.71	&25.78	&38.13	&40.12	&32.01	&16.86	&23.14	&22.64&	26.48&	70.00&	83.22&	31.75&	5.74	& 68.50&	53.50&	45.58	&37.82 \\
StreamingLLM (k=$0.22 n^{1/1000}$) & 23.37&	17.55&	33.93&	39.72&	29.63 &	20.57&	20.71&	22.93&	21.86&	68.00&	88.75&	38.34	&5.48&	69.12&	57.84&	53.42	&38.20 \\
StreamingLLM (k=1024) & 17.44	&8.68	&22.25	&35.37&	31.51&	15.97&	15.46&	20.06&	14.64&	38.00&	72.33&	29.10&	5.42	&69.50&	46.14&	45.09	&30.43 \\
StreamingLLM (k=$0.11 n^{1/1000}$) & 21.40	 & 16.92	 & 31.62	 & 38.45	 & 28.72 & 	18.59	 & 19.96	 & 22.49	 & 20.96 & 	66.50 & 	89.35 & 	38.43 & 	5.92 & 	69.00 & 	57.86 & 	51.8 & 	37.37\\

\bottomrule
\end{tabular}
}
\caption{Comparison of sparse attention (StreamingLLM, \cite{xtc+23}) with different window size strategies on length extrapolation benchmark including multiple datasets.}
% \vspace{-4mm}
\label{tab:main-table}
\end{table*}

\vspace{-2mm}
\section{Theoretical Insight V: Dynamic Top-\texorpdfstring{$k$}{} Strategy rather than Fixed One}\label{sec:dynamic_top_k}

In this section, we aim to develop the more advanced method for {\it stable sparse attention approximation} towards the efficient inference tasks of transformer-based models. We first give the theoretical analysis in Section~\ref{sub:insight_V_analysis} and the corresponding empirical evaluation in Section~\ref{sub:insight_V_evaluation}.

\subsection{Theoretical Analysis}\label{sub:insight_V_analysis}

We consider a dataset ${\cal D} := \{X^i\}_{i=1}^N$ with dataset size $N$. For all $i \in [N]$, we use $n_i$ to denote the context length, such that $X^i \in \R^{n_i \times d}$. Hence, the difference of the computational complexities of fixed Top-$k$ strategy and a dynamic (especially $O(n^C)$) strategy could be easily obtained in the Claim below.

\begin{claim}\label{clm:complexities}
    We have:
    \begin{itemize}
        \item {\bf Part 1.} Choosing the constant window size strategy $k = p$ for some constant integer $p > 0$. The computational complexity of a one-layer $p$-sparse attention to inference ${\cal D} = \{X^i\}_i^N$ is $\Theta(p \sum_{i=1}^N n_i)$.
        \item {\bf Part 2. } Choosing the constant window size strategy $k = \alpha \cdot n^C$ for some constant $\alpha, C > 0$. The computational complexity of a one-layer $p$-sparse attention to inference ${\cal D} = \{X^i\}_i^N$ is $\Theta(\alpha \sum_{i=1}^N n_i^{1+C})$.
    \end{itemize}
\end{claim}

\begin{proof}
    Recall the computational complexity of Definition~\ref{def:sparse_attn:informal} is $O(nk)$ for input $X \in \R^{n \times d}$ and $k$ window size. We then take the summation for each $X^i \in \R^{n_i \times d}$ in ${\cal D} := \{X^i\}_i^N$ to obtain the results of {\bf Part 1} and {\bf Part 2}.
\end{proof}

{\bf Remark. } We emphasize the goal of this section is not to present a novel method, in contrast, the design of strategy $k = \alpha n^C$ is very careless and retains a lot of room for improvement. However, the proposition we aim to sell in this section, which is: {\bf Using dynamic window size strategy rather than fixed window size strategy, the former is provably performs more efficient and higher performance than the latter}. We would like provide some simple but constructive insights to help better study the working mechanism behind and better design of sparse attention approximation. As we state below:
\begin{proposition}\label{pro:dynamic_top_k}
    For any window size strategy $k = p$ for some constant integer $p > 0$, there exist a context-length adaptive strategy $k = \alpha \cdot n^C$ for some constant $\alpha, C > 0$ that performs lower approximating error.
\end{proposition}

\begin{proof}
    Following Theorem~\ref{thm:approx_success:informal} and Theorem~\ref{thm:approx_failure:informal}, the conclusion of this proposition can be trivially proved by pluging suitbale choices of $\alpha$ and $C$.
\end{proof}

% {\bf Remark. } We emphasize the goal of this section is not to present a novel method, in contrast, the design of strategy $k = \alpha n^C$ is very careless and retains a lot of room for improvement. Here in this section, we aim to provide 

\subsection{Evaluation}\label{sub:insight_V_evaluation}

{\bf Setup.}
In this experiment, we apply $k=\alpha n^C$ window size strategy with choosing $C = 1/1000$ and $\alpha \in \{0.11, 0.22\}$. The baseline is set to the classical pruning-KV method, StreamingLLM \cite{xtc+23} with windows sizes $k \in \{1024, 2048\}$ that utilizes attention sink to relieve the bad approximating performance in efficient KV cache approaches. Furthermore, we use the pre-trained model LlaMa-3-8B-Instruct as the backbone of this experiment and we evaluate on the LongBench dataset \cite{blz+23}. Especially, each choice of $\alpha$ here performs less computional time compare to fixed strategies $\{1024, 2048\}$ on LongBench dataset, we thus conduct no additional experiments about computational efficiency in this setting.

{\bf Results.} The scores of each methods are shown in Table~\ref{tab:main-table} where a greater score represents a better performance for each sub-dataset. As demonstrating, the dynamic window size strategy showcases a strong approximating capability that compares with the fixed window size strategy under the settings $k=2048$ and $k=1024$. Besides, it also performs a robustness since $\alpha$ is scaled from $0.22$ to $0.11$, the average performance decreases a little. This confirms the correctness of Proposition~\ref{pro:dynamic_top_k} and we believe the potential of this insight for designing future {\it stable sparse attention approximation}.
\vspace{-2mm}

% \begin{claim}
%     Choosing the constant window size strategy $k = \alpha \cdot n^C$ for some constant $\alpha, C > 0$. The computational complexity of a one-layer $p$-sparse attention to inference ${\cal D} = \{X^i\}_i^N$ is $\Theta(\alpha \sum_{i=1}^N n_i^{1+C})$.
% \end{claim}

%% file: 8_conclusion.tex
\section{Conclusion and Future Discussion}

Prior fast algorithms for enhancing the efficiency of language models alleviate the quadratic time complexity of attention computation by assuming attention sparsity, whereas these algorithms lack an appropriate understanding of it. This work provides the first theoretical analysis of attention sparsity, furthermore, our theory leads to an obvious but important conclusion, that is {\it Attention is Provably, Naturally $n^C$-Sparse} for any constant $C \in (0, 1)$. 

In addition, this analysis of sparsity has brought us several brand-new insights. Here, we re-organize them as takeaways, we hope these can be some solid suggestion for improving future design of transformer-based architectures:

{\bf Takeaway 1: It's necessary to involve a high precision in attention computation to relieve attention collapse.} Since the proven concept of attention collapse, meanwhile, a high precision in attention computation means choosing $\epsilon$ with smaller value in Theorem~\ref{thm:attn_collapse:informal}, we thus suggest using super-high precision in attention computation to avoid forgetting in the case of long context.

{\bf Takeaway 2: A large constant window size for sparse attention approximation is enough, but a dynamic strategy will improve performance. } As the results shown in Table~\ref{tab:attn_approx_simulation} and Table~\ref{tab:main-table}, fixed window size strategy shows competitive approximating error. We suggest that if the model weights or dataset for inference is unknown to the model developers, using fixed window size strategy will be a insurance and effective method.

At the same time, we discuss two future direction as follows:

{\bf Future Direction 1: The upper bound on input and weight matrices (denotes $R$ in this paper) is vital to attention sparsity, a non-fixed weight, e.g. Rotary Positional Embedding (RoPE) \cite{sal+24} might help to address the long-term forgetting.} The conclusion of the former sentence is derived from Theorem~\ref{thm:attn_collapse:informal}. Hence a non-fixed weight (mathematically $R$ as Definition~\ref{def:R:informal}) leads to a converging lower bound on the number of ineffective entry in attention matrix. This connects to the long-term extrapolation problem of LLMs, which is a interesting future direction if we utilize the theory of this paper to explain some phenomenons.

{\bf Future Direction 2: Context-length and weight-norm adaptive efficient attention. } Theorem~\ref{thm:approx_success} shows the importance of context length and weight norm in {\it stable sparse attention approximation}. We would like to extend this framework onto general efficient attention computation, providing a novel perspective that is different with fine-grained complexity theory in attention \cite{as23_neurips}.

% \section*{Impact Statement}
% This paper presents work whose goal is to advance the field of Machine Learning. There are many potential societal consequences of our work, none of which we feel must be specifically highlighted here.

% one of which is the positive correlation between sparsity and attention weight norm R, directly solving the problem of prioritizing the application of fast algorithms in language models. Finally, the error analysis of sparse attention computation benefits from our theory, in which we simplified HyperAttention by introducing a scaled coefficient since the $(\epsilon, k)$-sparsity could be estimated, leads a clearer upper bound on error.

%% file: 9_app_more_related_work.tex
% \paragraph{Roadmap.}
% We organize the appendix as followed. In Section~\ref{sec:related_work_app}, we provide more works that are related to this paper. In Section~\ref{sec:prel_app} we give some more preliminaries. In Section~\ref{sec:kq_dist} we show that $KQ_i^\top$ is Gaussian distributed. In Section~\ref{sec:upper_bound_error_sparse} we show sparsity of attention with upper bound. In Section~\ref{sec:analysis} we provide analysis of our proposed algorithm. 

\section{More Related Work}\label{sec:related_work_app}
To complete the background of the related research, we provide some more related work in this section. 
The attention mechanism's computational intensity, particularly within transformer architectures, has been a significant focus, with \cite{dsxy23} delving into the theoretical aspects of data recovery via attention weights. Concurrently, the analysis of softmax attention's superiority by \cite{dsz23} and the innovative attention schemes inspired by softmax regression from \cite{dls23} contribute vital insights into attention mechanism optimization. The discourse extends to the computational frameworks of LLMs, where the theoretical examination of modern Hopfield models by \cite{hu2024computational} and the exploration of sparse Hopfield models by \cite{hu2023SparseHopfield} align closely with our research's objectives. Moreover, the innovative STanHop model by \cite{wu2023stanhop} and the outlier-efficient Hopfield layers presented by \cite{hu2024outlier} further underscore the burgeoning interest in enhancing computational efficiencies within neural architectures. In parallel, studies on various algorithmic optimizations, such as randomized and deterministic attention sparsification \cite{dms23_rand}, dynamic kernel sparsifiers \cite{djs+22}, and zero-th order algorithms for softmax attention optimization \cite{dlms23}, provide a comprehensive backdrop to our investigation into attention score sparsity. Further extending the discourse, studies by \cite{xu2024bishop}, \cite{hu2024nonparametric}, and Wu et al. \cite{wu2024uniform} delve into novel applications and optimizations of Hopfield networks, showcasing the diverse avenues for computational enhancement in neural network models.
\cite{csy24, csy23b} offer additional perspectives on optimizing LLMs, addressing copyright data protection and fine-tuning for unbiased in-context learning, which underscore the multifaceted challenges in LLM research. The work by \cite{sy23} on learning rate schedules further informs our understanding of model optimization and training efficiency.
Moreover, the diverse array of research spanning from Fourier circuits in neural networks \cite{gll+}, domain generalization \cite{smf+}, multitask finetuning \cite{xsw+}, to semi-supervised learning frameworks \cite{ssl, swl_neurips} and spectral analysis in class discovery \cite{ssll} offers a rich backdrop against which our work is situated. The Attention Sink phenomenon, discovered by \cite{xls+23, bnb24} and is helping to enhance LLMs in \cite{xtc+23, mil23}, is a distinct characteristic commonly observed in most attention networks of LLMs. This phenomenon is characterized by the disproportionate attention these models pay to a specific, often unrelated, token—usually the first token. These tokens are assigned high attention scores, indicating their significant influence within the network. These studies, along with \cite{scl+, dlg+, swl_iclr, ssl+} and \cite{qszz23, qrsxz22, qjs+22, gqsw22, qsz23, qsw23, qsy23} contributions in algorithmic optimizations and neural network analysis, not only broaden the landscape of our research but also underscore the dynamic and multifaceted nature of computational efficiency and theoretical analysis in the field of LLMs and beyond.

%% file: 10_app_preli.tex
\section{Preliminary}
\label{sec:prel_app}
% In Section~\ref{sec:notation_app} we provide some notations. In Section~\ref{sec:attn_def} we provide definitions related to attention computation. In Section~\ref{sec:input_def} we provide definitions of inputs. In Section~\ref{sec:problem_def} we provide the problem definition. In Section~\ref{sec:def_app} we provide some useful definitions. In Section~\ref{sec:facts_random_variable} we show some related facts about random variables. In Section~\ref{sec:softmax} we provide some fact for softmax computation. 

\subsection{Notations}
\label{sec:notation_app}
In this work, we use the following notations and definitions: 
\begin{itemize}
    \item For integer $n$, we use $[n]$ to denote the set $\{1, \dots, n\}$.
    \item We use $\mathbf{1}_n$ to denote all-$1$ vector in $\R^n$.
    \item The $\ell_p$ norm of a vector $x$ is denoted as $\| x \|_p$, for examples, $\| x \|_1 := \sum^n_{i=1} | x_i |$, $\| x \|_2 := ( \sum^n_{i=1} x_i^2 )^{1/2}$ and $\| x \|_\infty := \max_{i \in [n]} | x_i |$.
    \item For a vector $x \in \R^n$, $\exp(x) \in \R^n$ denotes a vector where whose i-th entry is $\exp(x_i)$ for all $i \in [n]$.
    \item For two vectors $x, y \in \R^n$, we denote $\langle x, y \rangle = \sum^n_{i=1}x_i y_i$ for $i \in [n]$.
    \item Given two vectors $x, y \in \R^n$, we denote $x \circ y$ as a vector whose i-th entry is $x_i y_i$ for all $i \in [n]$.
    \item For a vector $x \in \mathbb{R}^n$, $\diag(x) \in \mathbb{R}^{n \times n}$ is defined as a diagonal matrix with its diagonal entries given by $\diag(x)_{i,i} = x_i$ for $i = 1, ..., n$, and all off-diagonal entries are $0$.
    % \item We use $\E[x]$ to denote the expectation value of random variable $x \in \R^n$.
    % \item We use $\var[x]$ to denote the variance value of random variable $x \in \R^n$.
    \item We use $\mathrm{erf}(x)$ to denote the error function $\mathrm{erf}(x) = \frac{2}{\sqrt{\pi}}\int_{0}^{x} \exp(-t^2) \d t$, and $\mathrm{erf}^{-1}$ is denoted as the inverse function of $\mathrm{erf}(x)$.
    \item For any matrix $A \in \R^{m \times n}$, we use $A^\top$ to denote its transpose, we use $\|A\|_F$ to denote the Frobenius norm and $\|A\|_\infty$ to denote its infinity norm, i.e., $\|A\|_F := (\sum_{i \in [m]}\sum_{j \in [n]}A_{i,j}^2)^{1/2}$ and $\|A\|_\infty = \max_{i \in [m], j \in [n]}|A_{i,j}|$.
    \item For $\mu, \sigma \in \R$, we use $\N(\mu, \sigma^2)$ to denote Gaussian distribution with expectation of $\mu$ and variance of $\sigma^2$.
    \item For a mean vector $\mu \in \R^d$ and a covariance matrix $\Sigma \in \R^{d \times d}$, we use $\N(\mu, \Sigma^2)$ to denote the vector Gaussian distribution.
    \item We use $\E[\cdot]$ to denote the expectation and $\var[\cdot]$ to denote the variance. 
    \item We use $\Gamma(x)$ to denote the gamma function where $\Gamma(x) = \int_{0}^\infty t^{z-1} \exp(-t) \d t$.
    \item For an integer $k > 0$, we use $\chi_k^2$ to denote the Chi-squared distribution with $k$ degrees of freedom.
    \item Usually, we use $C \ge 1$ to denote a sufficient large constant.
\end{itemize}

\subsection{Basic Fact for Softmax}

\begin{fact}\label{fac:softmax_bias}
    For a vector $x \in \R^d$ and a scalar $b \in \R$, we have:
    \begin{align*}
        {\sf softmax}(x) = {\sf softmax}(x + b \cdot {\bf 1}_d)
    \end{align*}
\end{fact}

\subsection{Basic Facts for Calculation}

\begin{fact}\label{fac:sqrt_sum}
    For $a, b \ge 1$ and there exist a constant $C \ge 0$ such that
    \begin{align*}
        \sqrt{a} + \sqrt{b} \leq C \sqrt{a + b}
    \end{align*}
\end{fact}

\begin{fact}\label{fac:exp_sqrt_log}
    For a sufficient large $x \in \R$ ($x \ge 55$), we have
    \begin{align*}
        \exp(\sqrt{\log(x)}) \leq \sqrt{x}
    \end{align*}
\end{fact}

\subsection{Probability Tools}

Here, we state a probability toolkit in the following, including several helpful lemmas we'd like to use. Firstly, we provide the lemma about Chernoff bound in \cite{che52} below.

\begin{lemma}[Chernoff bound, \cite{che52}]\label{lem:chernoff}
    Let $X = \sum_{i=1}^n X_i$, where $X_i = 1$ with probability $p_i$ and $X_i = 0$ with probability $1 - p_i$, and all $X_i$ are independent. Let $\mu = \E[X] = \sum_{i=1}^n p_i$. Then
    \begin{itemize}
        \item $\Pr[X \geq (1 + \delta)\mu] \leq \exp(-\delta^2\mu/3)$, $\forall \delta >0$;
        \item $\Pr[X \leq (1 - \delta)\mu] \leq \exp(-\delta^2\mu/1)$, $\forall 0 < \delta < 1$.
    \end{itemize}
\end{lemma}

Next, we offer the lemma about Hoeffding bound as in \cite{hoe44}.

\begin{lemma}[Hoeffding bound, \cite{hoe44}]\label{lem:hoeffding}
    Let $X_1, \cdots , X_n$ denote $n$ independent bounded variables in $[a_i, b_i]$ for $a_i, b_i \in \R$. Let $X := \sum_{i=1}^n X_i$, then we have
    \begin{align*}
        \Pr[|X - \E[X]| \geq t] \leq 2\exp(- \frac{2t^2}{\sum_{i=1}^n (b_i -a_i)^2} )
    \end{align*}
\end{lemma}

We show the lemma of Bernstein inequality as \cite{ber24}.

\begin{lemma}[Bernstein inequality, \cite{ber24}]\label{lem:bernstein}
    Let $X_1, \cdots , X_n$ denote $n$ independent zero-mean random variables. Suppose $|X_i| \leq M$ almost surely for all $i$. Then, for all positive $t$,
    \begin{align*}
        \Pr[\sum_{i=1}^n X_i \geq t] \leq \exp(-\frac{t^2/2}{\sum_{j=1}^n \E[X_j^2] + Mt/3})
    \end{align*}
\end{lemma}

Then, we give the Khintchine’s inequality in \cite{khi23, haa81} as follows:

\begin{lemma}[Khintchine’s inequality, \cite{khi23, haa81}]\label{lem:khintchine}
    Let $\sigma_1, \cdots , \sigma_n$ be i.i.d sign random variables, and let $z_1 \cdots, z_n$ be real numbers. Then there are constants $C > 0$ so that for all $t > 0$
    \begin{align*}
        \Pr[|\sum_{i=1}^n z_i\sigma_i| \geq t\|z\|_2] \leq \exp(-Ct^2)
    \end{align*}
\end{lemma}

We give Hason-wright inequality from \cite{hw71, rv13} below.

\begin{lemma}[Hason-wright inequality, \cite{hw71, rv13}]\label{lem:hanson_wright}
    Let $x \in \R^n$ denote a random vector with independent entries $x_i$ with $\E[x_i] = 0$ and $|x_i|\leq K$ Let $A$ be an $n \times n$ matrix. Then, for every $t \geq 0$
    \begin{align*}
        \Pr[|x^\top Ax - \E[x^\top Ax]| > t] \leq 2\exp(-c \min\{ t^2/ (K^4 \|A\|_F^2), t/(K^2\|A\|)\})
    \end{align*}
\end{lemma}

We state Lemma 1 on page 1325 of Laurent and Massart \cite{lm00}.

\begin{lemma}[Lemma 1 on page 1325 of Laurent and Massart, \cite{lm00}]\label{lem:laurent_massart}
    Let $X \sim \mathcal{X}_k^2$ be a chi-squared distributed random variable with $k$ degrees of freedom. Each one has zero mean and $\sigma^2$ variance. Then
    \begin{align*}
        \Pr[X - k\sigma^2 \geq (2\sqrt{kt}+2t)\sigma^2] \leq \exp(-t) & ~ \\
        \Pr[X - k\sigma^2 \geq 2\sqrt{kt}\sigma^2] \leq \exp(-t) & ~
    \end{align*}
\end{lemma}

Here, we provide a tail bound for sub-exponential distribution \cite{fkz11}.

\begin{lemma}[Tail bound for sub-exponential distribution, \cite{fkz11}]
    We say $X \in \mathrm{SE}(\sigma^2, \alpha)$ with parameters $\sigma > 0$, $\alpha > 0$, if
    \begin{align*}
        \E[e^{\lambda X}] \leq \exp(\lambda^2 \sigma^2/2), \forall | \lambda | < 1 / \alpha.
    \end{align*}
    Let $X \in \mathrm{SE}(\sigma^2, \alpha)$ and $\E[X] = \mu$, then:
    \begin{align*}
        \Pr[|X- \mu|\geq t] \leq \exp(-0.5\min\{t^2/\sigma^2, t/\alpha\})
    \end{align*}
\end{lemma}

In the following, we show the helpful lemma of matrix Chernoff bound as in \cite{tro11, ldfu23}.

\begin{lemma}[Matrix Chernoff bound, \cite{tro11, ldfu23}]
    Let $\mathcal{X}$ be a finite set of positive-semidefinite matrices with dimension $d \times d$, and suppose that
    \begin{align*}
        \max_{X \in \mathcal{X}} \lambda_{\rm max}(X)\leq B.
    \end{align*}
    Sample $\{X_1, \cdots , X_n\}$ uniformly at random from $\mathcal{X}$ without replacement. We define $\mu_{\rm min}$ and $\mu_{\rm max}$ as follows:
    \begin{align*}
        \mu_{\rm min} := n \cdot \lambda_{\rm min}(\E_{X \in \mathcal{X}}(X)) & ~ \\
        \mu_{\rm max} := n \cdot \lambda_{\rm max}(\E_{X \in \mathcal{X}}(X)). & ~
    \end{align*}
    Then
    \begin{align*}
        & ~\Pr[\lambda_{\rm min}(\sum_{i=1}^n X_i) \leq (1 - \delta) \mu_{\rm min}] \leq d \cdot \exp(-\delta^2\mu_{\rm min}/B) \text{~for~} \delta \in (0, 1], \\
        & ~\Pr[\lambda_{\rm max}(\sum_{i=1}^n X_i) \geq (1 + \delta) \mu_{\rm max}] \leq d \cdot \exp(-\delta^2\mu_{\rm max}/(4B)) \text{~for~} \delta \geq  0.
    \end{align*}
\end{lemma}

%% file: 12_app_defs.tex
\section{Problem Definitions}

\subsection{Input Assumption}

\begin{definition}\label{def:X}
    We consider for any input matrix to an attention network $X \in\R^{n \times d}$ where integer $n$ denotes the input length and $d$ denotes the dimension. We assume:
    \begin{itemize}
        \item {\bf Independent Entries.} For any two entries $X_{i_1, j_1}$ and $X_{i_2, j_2}$ in matrix $X$, $\forall i_1, i_2 \in [n]$ and $j_1, j_2 \in [d]$, they are independent.
        \item {\bf Bounded Entries.} For failure probability $\delta \in (0, 0.1)$. With a probability $1 - \delta$, the entry $X_{i, j}$ in matrix $X$, $\forall i \in [n]$ and $j\in [d]$, we have $| X_{i, j}  | \leq B$ for some positive constant $B > 0$.
    \end{itemize}
\end{definition}

\subsection{Attention Computation}
\label{sec:attn_def}
% Here we introduce the following definitions of attention with layer normalization. 
% \begin{definition}[Layer normalization]\label{def:Ln}
%     If the following conditions hold:
%     \begin{itemize}
%         \item Given an input matrix $X \in \R^{n \times d}$, denote the $i$-th row of $X$ as $X_i \in \R^d$ for any $i \in [n]$.
%         \item Let $\gamma, \beta \in \R^d$ be denoted as the weight and bias of layer normalization.
%         \item $\E[X_i] = \frac{1}{d}\sum_{j=1}^d X_{i, j} \in \R^d$ denotes the mean value of entries of $X_i$.
%         \item $\var[X_i] = \frac{1}{d}\sum_{j=1}^d (X_{i, j} - \E[X_i])^2 \in \R$ denotes the variance of entries of $X_i$.
%     \end{itemize}
%     We define the layer normalization of input $X \in \R^{n \times d}$ as ${\sf Ln}(X) \in \R^{n \times d}$, the $i$-th row for $i \in [n]$ is given by:
%     \begin{align*}
%         {\sf Ln}_i(X) := \frac{X_i - \E[X_i] \cdot {\bf 1}_d}{\sqrt{\var[X_i]}} \circ \gamma + \beta \in \R^d
%     \end{align*}
% \end{definition}

% \begin{remark}\label{rem:X}
%     The input matrix $X$ in Definition~\ref{def:Ln} is the input before Layer normalization in the transformer, to distinguish, we denote the input {before} Layer normalization as $X_{\rm pre} \in \R^{n \times d}$. We define:
%     \begin{align*}
%         X_i := \frac{X_{{\rm pre}, i} - \E[X_{{\rm pre}, i}] \cdot {\bf 1}_d}{\sqrt{\var[X_{{\rm pre}, i}]}} \in \R^d.
%     \end{align*}
%     Then, we can rewrite
%     \begin{align*}
%         {\sf Ln}_i(X) = X_i \circ \gamma + \beta \in \R^d.
%     \end{align*}
% \end{remark}

\begin{definition}[Attention computation]\label{def:attn}
    If the following conditions hold:
    \begin{itemize}
        \item Let $W_Q, W_K, W_V \in \R^{d \times d}$ be denoted as Query, Key and Value projection matrices of attention.
        \item Given an input $X \in \R^{n \times d}$ that holds properties in Definition~\ref{def:X}.
        \item Define Query, Key and Value states matrices $Q := XW_Q, K:= XW_K, V:= XW_V \in \R^{n \times d}$.
        \item $A  := \exp(Q K^\top / \sqrt{d}) \in \R^{n \times n}$.
        \item $D := \diag(A{\bf 1}_n) \in \R^{n \times n}$.
    \end{itemize}
    Then we have attention computation ${\sf Attn}(Q, K, V) \in \R^{n \times d}$ as follows:
    \begin{align*}
        {\sf Attn}(Q, K, V) := D^{-1} A V
    \end{align*}
\end{definition}

\subsection{\texorpdfstring{$\epsilon$}{}-Approximated \texorpdfstring{$k$}{}-Sparse Softmax Vector}
\label{sec:problem_def}

In order to describe the sparsity of the softmax, we define the following notation. 

\begin{definition}\label{def:S}
    For a vector $u \in \R^n$ and $\epsilon \ge 0$, we define sparse set $\mathcal{S}_\epsilon(u)$ as follows:
    \begin{align*}
        \mathcal{S}_{\epsilon}(u) := \Big\{ i \in [n] ~\Big|~ |u_i| \le \epsilon \Big\}
    \end{align*}
\end{definition}

\begin{definition}[$(\epsilon, k)$-sparsity]
    For a vector $u \in \R^{n}$, we say $u$ is $(\epsilon, k)$-sparse if for a constant $\epsilon \in (0, 1)$, it holds that
    \begin{align*}
        | \mathcal{S}_{\epsilon}(u) | \ge n-k.
    \end{align*}
\end{definition}

% \begin{definition}\label{def:P_sparse}
%     Let $\epsilon \in (0, 1)$ be a given error parameter, and integer $k \in [0, n]$ be a sparsity parameter. Let $D \in \R^{n \times n}$ and $A \in \R^{n \times n}$ be defined as Definition~\ref{def:attn}. For any $i \in [n]$, we define
%     \begin{align*}
%         P_{\it sparse}(\epsilon, k)
%     := \Pr[D_{i, i}^{-1} A_i \text{~is~} (\epsilon, k)\text{-sparse}]. 
%     \end{align*}
%     % \Chiwun{The $P_{\it sparse}(\epsilon, k)$ is complicated to define, can you help me to re-write it?}\Yichuan{Improved}
% \end{definition}

\subsection{Sparse Attention}

\begin{definition}\label{def:T}
    For a vector $u \in \R^{n}$. Given a sparsity integer $k$. Let ${\cal S}_\epsilon$ be defined as Definition~\ref{def:S} for some error $\epsilon > 0$.
    We define the top-$k$ set ${\cal T}_{k}(u) := \{ i \in [n] ~|~ {\cal S}_{u_i}(u) \ge n - k \}$.
\end{definition}

\begin{definition}\label{def:topk}
    If the following conditions hold:
    \begin{itemize}
        \item For a vector $u \in \R^{n}$.
        \item Given a sparsity integer $k$.
        \item Let ${\cal S}_\epsilon$ be defined as Definition~\ref{def:S} for some error $\epsilon > 0$.
        \item Denote a top-$k$ set ${\cal T}_{k}(u) := \{ i \in [n] ~|~ {\cal S}_{u_i}(u) \ge n - k \}$ as Definition~\ref{def:T}.
    \end{itemize}
    Then we define
    \begin{align*}
        {\sf topk}(u) := [ u_i \cdot {\bf 1}_{i \in {\cal T}_{k}(u)} ]_{i\in [n]} \in \R^n
    \end{align*}
\end{definition}

We define the sparse attention computation as follows:
\begin{definition}\label{def:sparse_attn}
    If the following conditions hold:
    \begin{itemize}
        \item Let $W_Q, W_K, W_V \in \R^{d \times d}$ be denoted as Query, Key and Value projection matrices of attention.
        \item Given an input $X \in \R^{n \times d}$ that holds properties in Definition~\ref{def:X}.
        % \item Scalar $B \ge 0$.
        % \item Let $|X_{i, j}| \leq B$ as Assumption~\ref{ass:bounded_X} for $i \in [n]$, $j \in [d]$.
        % \item Let ${\sf Ln}(X) \in \R^{n \times d}$ be defined as Definition~\ref{def:Ln}, where the $i$-th row of ${\sf Ln}(X)$ is ${\sf Ln}_i(X)$ for $i \in [n]$.
        % \item Let $\gamma, \beta \in \R^d$ be defined as Definition~\ref{def:Ln}.
        % \item Denote $Z := {\sf Ln}(X) \in \R^{n \times d}$, denote the $i$-th row of $X$ as $Z_i \in \R^d$ for any $i \in [n]$.
        \item Define Query, Key and Value states matrices $Q := XZW_Q, K:= XW_K, V:= XW_V \in \R^{n \times d}$.
        \item $A  := \exp(Q K^\top / \sqrt{d}) \in \R^{n \times n}$.
        \item Let ${\sf topk}$ be defined as Definition~\ref{def:topk}.
        \item Define $A_{\rm spar} := \begin{bmatrix}
        {\sf topk}(A_{1}), & {\sf topk}(A_{2}), & \cdots, & {\sf topk}(A_{n})
    \end{bmatrix}^\top \in \R^{n \times n}$.
        % \item Denote the scale coefficient $\alpha \in [1, \frac{1}{(n-k)\epsilon}]$.
        \item $D_{\rm spar} :=  \diag(A_{\rm spar} {\bf 1}_n)\in \R^{n \times n}$.
        \item $\delta \in (0, 0.1)$.
        \item Let $R \ge 0$ be defined as Definition~\ref{def:R}.
    \end{itemize}
    The sparse attention computation ${\sf SparseAttn}(Q, K, V) \in \R^{n \times d}$ is given by:
    \begin{align*}
        {\sf SparseAttn}(Q, K, V) := D_{\rm spar}^{-1} A_{\rm spar} V \in \R^{n \times d}
    \end{align*}
\end{definition}

\subsection{Helpful Definitions}
\label{sec:def_app}
We introduce the following algebraic lemmas to be used later. 
% \begin{definition}\label{def:B}
%     Let $\beta \in \R^{d}$ be defined as Definition~\ref{def:Ln}. We define $B \in \R^{n \times d}$ that for $i \in [n]$, its $i$-th row is defined as follows: 
%     \begin{align*}
%         B_i := \beta
%     \end{align*}
% \end{definition}

\begin{definition}\label{def:W}
    Let Query and Key projection matrices $W_Q, W_K \in \R^{d \times d}$ be defined as Definition~\ref{def:attn}. We define
    \begin{align*}
        W := W_Q W_K^\top / \sqrt{d}.
    \end{align*}
    % Then, we can rewrite
    % \begin{align*}
    %     QK^\top = Z W Z^\top
    % \end{align*}
\end{definition}

\begin{definition}\label{def:R}
    If the following conditions hold:
    \begin{itemize}
        % \item Given an input $X \in \R^{n \times d}$ that holds properties in Definition~\ref{def:X}.
        \item Let $W \in \R^{d \times d}$ be define as Definition~\ref{def:W}.
        % \item Denote $\gamma_{\rm max} = \max_{j \in [d]} \{ \beta_j, \gamma_j \}$.
    \end{itemize}
    Then for any $i \in [n]$, we define:
    \begin{align*}
        R := B^2 \| W\|_F.
    \end{align*}
\end{definition}

% \begin{definition}
% \label{def:W*}
%     Let $\gamma \in \R^d$ be defined as Definition~\ref{def:Ln} and $W \in \R^{d \times d}$ be defined as Definition~\ref{def:W}. We define the matrix
%     \begin{align*}
%         W_* := \diag(\gamma) W \diag(\gamma)
%     \end{align*}
% \end{definition}

%% file: 13_app_attention_sparsity.tex
\section{Attention Sparsity}\label{sec:main_result}

\subsection{Main Result 1: Attention Sparsity with Upper Bound on Error}

\begin{theorem}\label{thm:main_result}
    If the following conditions hold:
    \begin{itemize}
        \item Let $W_Q, W_K, W_V \in \R^{d \times d}$ be denoted as Query, Key and Value projection matrices of attention.
        \item Given an input $X \in \R^{n \times d}$ that holds properties in Definition~\ref{def:X}.
        % \item Let $|X_{i, j}| \leq B$ as Assumption~\ref{ass:bounded_X} for $i \in [n]$, $j \in [d]$.
        % \item Let ${\sf Ln}(X) \in \R^{n \times d}$ be defined as Definition~\ref{def:Ln}, where the $i$-th row of ${\sf Ln}(X)$ is ${\sf Ln}_i(X)$ for $i \in [n]$.
        % \item Let $\gamma, \beta \in \R^d$ be defined as Definition~\ref{def:Ln}.
        % \item Denote $Z := {\sf Ln}(X) \in \R^{n \times d}$.
        \item Define Query, Key and Value states matrices $Q := XW_Q, K:= XW_K, V:= XW_V \in \R^{n \times d}$.
        \item $A  := \exp(Q K^\top / \sqrt{d}) \in \R^{n \times n}$.
        \item $D := \diag(A{\bf 1}_n) \in \R^{n \times n}$.
        \item Denote $\beta_{i} := Bd \max_{j_1 \in [d]} |\E[\sum_{j_2=1}^d W_{j_1, j_2} \cdot X_{i, j_2}] |$.
        \item Define $\Gamma := [ \beta_1 \cdot {\bf 1}_n, \beta_2 \cdot {\bf 1}_n, \cdots, \beta_n  \cdot {\bf 1}_n ]^\top \in \R^{n \times n}$
        \item $\wt{A}  := \exp(Q K^\top / \sqrt{d} + \Gamma) \in \R^{n \times n}$.
        \item $\wt{D} := \diag(\wt{A} {\bf 1}_n) \in \R^{n \times n}$.
        \item $\delta \in (0, 0.1)$.
        \item Let $R \ge 0$ be defined as Definition~\ref{def:R}.
        \item Given sparsity integer $k \le n$.
        \item Denote $T := \exp(\sqrt{\log(n(n-k)d/\delta)})$.
        \item Let $\mathcal{S}_\epsilon$ be defined as Definition~\ref{def:S}.
    \end{itemize}
    If we choose $\epsilon \ge \frac{T^{O(R)}}{n})$,
    then with a probability at least $1 - \delta$, we have
    \begin{align*}
        \Big| \mathcal{S}_\epsilon(D^{-1}_{i_1, i_1} A_{i_1}) \Big| = \Big| \mathcal{S}_\epsilon(\wt{D}^{-1}_{i_1, i_1} \wt{A}_{i_1}) \Big| \ge n - k
    \end{align*}
\end{theorem}

\begin{proof}
    {\bf Remark.} We re-denote $\mathcal{S}_\epsilon = \mathcal{S}_\epsilon( D_{i_1, i_1}^{-1} A_{i_1, *} )$ in the statement, and $i_2 \in \mathcal{S}_\epsilon$.

    Following Part 1 of Lemma~\ref{lem:bound_D}, with a probability at least $1 - \delta_1$, we have
    \begin{align}\label{eq:bound_A}
        \wt{A}_{i_1, i_2} \leq \exp(O(R) \cdot \sqrt{\log((n-k)d/\delta_1)})
    \end{align}

    Following Part 3 of Lemma~\ref{lem:bound_D}, with a probability at least $1 - \delta_2$, we have
    \begin{align}\label{eq:bound_D}
        \wt{D}_{i_1, i_1}^{-1} \leq \exp(O(R) \cdot \sqrt{\log(nd/\delta_2)}) / n
    \end{align}

    Now we combine Eq.~\eqref{eq:bound_A} and Eq.~\eqref{eq:bound_D}, with a probability at least $1 - \delta_1 - \delta_2$, we have
    \begin{align*}
        D_{i_1, i_1}^{-1} A_{i_1, i_2}
        = & ~ \wt{D}_{i_1, i_1}^{-1} \wt{A}_{i_1, i_2} \\
        \leq & ~ \exp(O(R) \cdot \sqrt{\log((n-k)d/\delta_1)} + O(R) \cdot  \sqrt{\log(nd/\delta_2)}) / n \\
        \leq & ~ \exp(O(R) \cdot \sqrt{\log((n-k)d/\delta_1) + \log(nd/\delta_2)}) / n \\
        \leq & ~ \exp(O(R) \cdot \sqrt{\log(n(n-k)d^2/(\delta_1\delta_2))}) / n \\
        \leq & ~ \exp( O(R) \cdot \sqrt{\log(n(n-k)d/\delta)}) / n \\
        \leq & ~ \exp( \sqrt{\log(n(n-k)/\delta)})^{O(R)} / n \\
        \leq & ~ T^{O(R)} \cdot n^{-1}
    \end{align*}
    where the first step follows from Fact~\ref{fac:softmax_bias}, the second step follows from Eq.~\eqref{eq:bound_A} and Eq.~\eqref{eq:bound_D}, the third step follows from Fact~\ref{fac:sqrt_sum}, the fourth step follows from simple algebras, the fifth step follows from choosing $\delta_1 = \delta_2 = \delta / 2$, the sixth step follows from simple algebras, the last step follows from the definition of $T$.
\end{proof}

\subsection{Main Result 2: Attention Collapse}

\begin{theorem}\label{thm:attn_collapse}
    If the following conditions hold:
    \begin{itemize}
        \item Let $W_Q, W_K, W_V \in \R^{d \times d}$ be denoted as Query, Key and Value projection matrices of attention.
        \item Given an input $X \in \R^{n \times d}$ that holds properties in Definition~\ref{def:X}.
        % \item Let $|X_{i, j}| \leq B$ as Assumption~\ref{ass:bounded_X} for $i \in [n]$, $j \in [d]$.
        % \item Let ${\sf Ln}(X) \in \R^{n \times d}$ be defined as Definition~\ref{def:Ln}, where the $i$-th row of ${\sf Ln}(X)$ is ${\sf Ln}_i(X)$ for $i \in [n]$.
        % \item Let $\gamma, \beta \in \R^d$ be defined as Definition~\ref{def:Ln}.
        % \item Denote $Z := {\sf Ln}(X) \in \R^{n \times d}$.
        \item Define Query, Key and Value states matrices $Q := XW_Q, K:= XW_K, V:= XW_V \in \R^{n \times d}$.
        \item $A  := \exp(Q K^\top / \sqrt{d}) \in \R^{n \times n}$.
        \item $D := \diag(A{\bf 1}_n) \in \R^{n \times n}$.
        \item Denote $\beta_{i} := Bd \max_{j_1 \in [d]} |\E[\sum_{j_2=1}^d W_{j_1, j_2} \cdot X_{i, j_2}] |$.
        \item Define $\Gamma := [ \beta_1 \cdot {\bf 1}_n, \beta_2 \cdot {\bf 1}_n, \cdots, \beta_n  \cdot {\bf 1}_n ]^\top \in \R^{n \times n}$
        \item $\wt{A}  := \exp(Q K^\top / \sqrt{d} + \Gamma) \in \R^{n \times n}$.
        \item $\wt{D} := \diag(\wt{A} {\bf 1}_n) \in \R^{n \times n}$.
        \item $\delta \in (0, 0.1)$.
        \item Let $R \ge 0$ be defined as Definition~\ref{def:R}.
        \item Assuming $R = o(\sqrt{\log(n)})$.
        \item Given sparsity integer $k \le n$.
        \item Denote $T := \exp(\sqrt{\log(n(n-k)d/\delta)})$.
        \item Let $\mathcal{S}_\epsilon$ be defined as Definition~\ref{def:S}.
    \end{itemize}
    For any $\epsilon > 0$, with probability at least $1 - \delta$, we have:
    \begin{align*}
        \lim_{n \rightarrow +\infty} | \mathcal{S}_\epsilon( D_{i, i}^{-1} A_i ) | = n - 1
    \end{align*}
\end{theorem}

\begin{proof}
    In order to choose $k$ that meets the $\epsilon$-approximated sparsity, we have:
    \begin{align*}
        \epsilon \ge \exp\Big( O(R) \cdot \sqrt{ \log(n \cdot (n-k)d/\delta) } \Big)
    \end{align*}
    where this step follows from Theorem~\ref{thm:main_result}.

    We obtain:
    \begin{align}\label{eq:bound_k}
        k \le & ~ n - \exp\Big( O(\frac{\log^2(\epsilon \cdot n)}{R^2}) \Big) \cdot \frac{\delta}{nd}
    \end{align}

    Hence, we have:
    \begin{align*}
        \lim_{n \rightarrow +\infty} | \mathcal{S}_\epsilon( D_{i, i}^{-1} A_i ) |
        \ge & ~ \lim_{n \rightarrow +\infty} (n - k) \\
        \ge & ~ \lim_{n \rightarrow +\infty} \exp\Big( O(\frac{\log^2(\epsilon \cdot n)}{R^2}) \Big) \cdot \frac{\delta}{nd} \\
        = & ~ n-1
    \end{align*}
    where the first step follows from Definition~\ref{def:S}, the second step follows from Eq.\eqref{eq:bound_k}, the last step follows from $R = o(\sqrt{\log(n)})$ and $\langle D_{i, i}^{-1} A_i, {\bf 1}_n \rangle =1$, then 
    \begin{align*}
        \max | \mathcal{S}_\epsilon( D_{i, i}^{-1} A_i ) | = n - 1.
    \end{align*}
\end{proof}

\subsection{Bounding \texorpdfstring{$D^{-1} $}{}}

\begin{lemma}\label{lem:bound_D}
    If the following conditions hold:
    \begin{itemize}
        \item Let $W_Q, W_K, W_V \in \R^{d \times d}$ be denoted as Query, Key and Value projection matrices of attention.
        \item Given an input $X \in \R^{n \times d}$ that holds properties in Definition~\ref{def:X}.
        % \item Scalar $B \ge 0$.
        % \item Let $|X_{i, j}| \leq B$ as Assumption~\ref{ass:bounded_X} for $i \in [n]$, $j \in [d]$.
        % \item Let ${\sf Ln}(X) \in \R^{n \times d}$ be defined as Definition~\ref{def:Ln}, where the $i$-th row of ${\sf Ln}(X)$ is ${\sf Ln}_i(X)$ for $i \in [n]$.
        % \item Let $\gamma, \beta \in \R^d$ be defined as Definition~\ref{def:Ln}.
        % \item Denote $Z := {\sf Ln}(X) \in \R^{n \times d}$, denote the $i$-th row of $X$ as $Z_i \in \R^d$ for any $i \in [n]$.
        \item Define Query, Key and Value states matrices $Q := XW_Q, K:= XW_K, V:= XW_V \in \R^{n \times d}$.
        \item Denote $\beta_{i} := Bd \max_{j_1 \in [d]} |\E[\sum_{j_2=1}^d W_{j_1, j_2} \cdot X_{i, j_2}] |$.
        \item Define $\Gamma := [ \beta_1 \cdot {\bf 1}_n, \beta_2 \cdot {\bf 1}_n, \cdots, \beta_n  \cdot {\bf 1}_n ]^\top \in \R^{n \times n}$
        \item $\wt{A}  := \exp(Q K^\top / \sqrt{d} + \Gamma) \in \R^{n \times n}$.
        \item $\wt{D} := \diag(\wt{A} {\bf 1}_n) \in \R^{n \times n}$.
        % \item Let $W \in \R^{d \times d}$ be define as Definition~\ref{def:W}.
        % \item Denote $\gamma_{\rm max} = \max_{j \in [d]} \{ \beta_j, \gamma_j \}$.
        % \item Denote $C \ge 8$ as a sufficient large constant.
        \item $\delta \in (0, 0.1)$.
        \item Let $R \ge 0$ be defined as Definition~\ref{def:R}.
    \end{itemize}
    Then with a probability at least $1 - \delta$, we have
    \begin{itemize}
        \item Part 1. For $i_1, i_2 \in [n]$
        \begin{align*}
            \exp(- O(R) \cdot \sqrt{ \log(d/\delta) } ) \leq \wt{A}_{i_1, i_2} \leq \exp( O(R) \cdot \sqrt{ \log(d/\delta) } )
        \end{align*}
        \item Part 2. For $i_1 \in [n]$
        \begin{align*}
            n \cdot \exp(- O(R) \cdot \sqrt{ \log(nd/\delta) } ) \leq \wt{D}_{i_1, i_1} \leq n \cdot \exp( O(R) \cdot \sqrt{ \log(nd/\delta) } )
        \end{align*}
        \item Part 2. For $i_1 \in [n]$
        \begin{align*}
            \wt{D}_{i_1, i_1}^{-1} \leq \exp( O(R) \cdot \sqrt{\log(nd/\delta)}) / n
        \end{align*}
    \end{itemize}
\end{lemma}

\begin{proof}
    {\bf Proof of Part 1.}
    We have
    \begin{align*}
        |\wt{A}_{i_1, i_2}| = & ~ \exp((QK^\top)_{i_1, i_2}) \\
        \leq &  ~ \exp( O(R) \cdot \sqrt{ \log(d/\delta) } )
    \end{align*}
    where the first step follows from the definition of $A$, the second step follows from Lemma~\ref{lem:QK_concen}.

    {\bf Proof of Part 2.}
    This proof follows from the union bound of Part 1 of this Lemma and the Definition of $D$.

    {\bf Proof of Part 3.}
    This proof follows from the lower bound on $D_{i_1, i_1}$ and simple algebras.
\end{proof}

\subsection{Concentration on \texorpdfstring{$QK^\top$}{}}

\begin{lemma}\label{lem:QK_concen}
    If the following conditions hold:
    \begin{itemize}
        \item Let $W_Q, W_K, W_V \in \R^{d \times d}$ be denoted as Query, Key and Value projection matrices of attention.
        \item Given an input $X \in \R^{n \times d}$ that holds properties in Definition~\ref{def:X}.
        % \item Let ${\sf Ln}(X) \in \R^{n \times d}$ be defined as Definition~\ref{def:Ln}, where the $i$-th row of ${\sf Ln}(X)$ is ${\sf Ln}_i(X)$ for $i \in [n]$.
        % \item Let $\gamma, \beta \in \R^d$ be defined as Definition~\ref{def:Ln}.
        % \item Denote $Z := {\sf Ln}(X) \in \R^{n \times d}$, denote the $i$-th row of $Z$ as $Z_i \in \R^d$ for any $i \in [n]$.
        \item Define Query, Key and Value states matrices $Q := XW_Q, K:= XW_K, V:= XW_V \in \R^{n \times d}$.
        \item Let $W \in \R^{d \times d}$ be define as Definition~\ref{def:W}.
        % \item Denote $\gamma_{\rm max} = \max_{j \in [d]} \{ \beta_j, \gamma_j \}$.
        \item Denote $C \ge 1$ as a sufficient large constant.
        \item $\delta \in (0, 0.1)$.
        \item Let $R \ge 0$ be defined as Definition~\ref{def:R}.
        \item For $i_1, i_2 \in [n]$
    \end{itemize}
    Then with a probability at least $1 - \delta$, we have
    \begin{align*}
        | (QK^\top)_{i_1, i_2} | \leq O(R) \cdot \sqrt{\log(d / \delta)} + \beta_{i_2}
    \end{align*}
\end{lemma}

\begin{proof}
    We have:
    \begin{align*}
        | (QK^\top)_{i_1, i_2} |
        = & ~ | (XW_Q W_K^\top X^\top / \sqrt{d})_{i_1, i_2} | \\
        = & ~ | (X W X^\top)_{i_1, i_2} | \\
        = & ~ | X_{i_1}^\top W X_{i_2} | \\
        = & ~ | \sum_{j_1=1}^d \sum_{j_2=1}^d W_{j_1, j_2} \cdot X_{i_1, j_1} X_{i_2, j_2} | \\
        \leq & ~ B | \sum_{j_1=1}^d \sum_{j_2=1}^d W_{j_1, j_2} \cdot X_{i_2, j_2} |
        % = & ~ | X_{i_1}^\top W (X_{i_2} - \E[X_{i_2}]) + X_{i_1}^\top W\E[X_{i_2}]  | \\
        % \leq & ~ | X_{i_1}^\top W (X_{i_2} - \E[X_{i_2}]) | + |X_{i_1}^\top W\E[X_{i_2}]  | \\
        % \leq & ~ | X_{i_1}^\top W (X_{i_2} - \E[X_{i_2}]) | + \beta_{i_1, i_2} \\
        % = & ~ |  \sum_{j_1=1}^d \sum_{j_2=1}^d W_{j_1, j_2} \cdot X_{i_1, j_2} \cdot (X_{i_2, j_2} - \E[X_{i_2, j_2}]) | + \beta_{i_1, i_2} \\
        % \leq & ~ (M + B) |  \sum_{j_1=1}^d \sum_{j_2=1}^d W_{j_1, j_2} \cdot (X_{i_2, j_2} - \E[X_{i_2, j_2}]) | + \beta_{i_1, i_2} \\
    \end{align*}
    where the first step follows from $Q := XW_Q, K:= XW_K$, the second step follows from Definition~\ref{def:R}, the third and fourth steps follow from simple algebras, the fifth step follows from Definition~\ref{def:X}.

    We then apply Hoeffding inequality (Lemma~\ref{lem:hoeffding}) to each $W_{j_1, j_2} \cdot X_{i_2, j_2}$. Hence, with a probability at least $1 - \delta$, we have:
    \begin{align*}
        | \sum_{j_2=1}^d W_{j_1, j_2} \cdot X_{i_2, j_2} - \E[\sum_{j_2=1}^d W_{j_1, j_2} \cdot X_{i_2, j_2}] | 
        \leq & ~  O(B) \cdot \| W_{j_1} \|_2 \cdot \sqrt{\log(d / \delta)}
    \end{align*}
    since $|X_{i_2, j_2} | \leq B$ for any $j_2 \in [d]$.

    By triangle inequality, we have:
    \begin{align}\label{eq:concen_wx}
        | \sum_{j_2=1}^d W_{j_1, j_2} \cdot X_{i_2, j_2} | 
        \leq & ~  O(B) \cdot \| W_{j_1} \|_2 \cdot \sqrt{\log(d / \delta)} +  |\E[\sum_{j_2=1}^d W_{j_1, j_2} \cdot X_{i_2, j_2}] |
    \end{align}

    We obtain:
    \begin{align*}
        B | \sum_{j_1=1}^d \sum_{j_2=1}^d W_{j_1, j_2} \cdot X_{i_2, j_2} |
        \leq & ~  O(B^2) \cdot \| W \|_F \cdot \sqrt{\log(d / \delta)} +Bd \max_{j_1 \in [d]} |\E[\sum_{j_2=1}^d W_{j_1, j_2} \cdot X_{i_2, j_2}] | \\
        = & ~ O(R) \cdot \sqrt{\log(d / \delta)} + \beta_{i_2}
    \end{align*}
    where the first step follows from Eq~\ref{eq:concen_wx} and Fact~\ref{fac:sqrt_sum} ($\|W\|_F = \sum_{j_1=1}^d \|W_{j_1}\|_2^2$), the second step follows from Definition~\ref{def:R} and define
    \begin{align*}
        \beta_{i_2} := Bd \max_{j_1 \in [d]} |\E[\sum_{j_2=1}^d W_{j_1, j_2} \cdot X_{i_2, j_2}] |
    \end{align*}
\end{proof}

\begin{remark}\label{rem:addtional_term}
    The formal results of Lemma~\ref{lem:QK_concen} in the appendix have slight differences with the informal forms, in which we omit the additional terms of each upper bound since such terms are trivially some constants. Fact~\ref{fac:softmax_bias} shows any constant bias term added before the softmax function will not change the output. We thus simplify the equations for tighter boundaries and more convenient notation.
\end{remark}

%% file: 14_error_analysis.tex
\section{Sparse Attention Approximation}

% \subsection{Definitions}

% \subsection{General Conditions for Approximation}

% \begin{definition}[Properties]\label{def:general_conditions}
%     We state the following properties:
%     \begin{itemize}
%         \item General Conditions 1. $B = O\Big(\log^{\frac{1}{4}}(n)\Big)$.
%         \item General Conditions 2. $d = O(\log(n))$.
%         \item General Conditions 3. $\|W\|_2 = o(1)$.
%         \item General Conditions 4. $\gamma_{\rm max} = O(1/\sqrt{d})$.
%         \item General Conditions 5. $R = C_a \sqrt{\log(n)}$ for constant $C_a \in (0, 0.001)$.
%         \item General Conditions 6. $\|V\|_\infty = O(\sqrt{\log(n)})$.
%     \end{itemize}
% \end{definition}

\subsection{Main Result 3: Upper Bound on Error}

\begin{theorem}\label{thm:approx_success}
    If the following conditions hold:
    \begin{itemize}
        \item Let Query, Key and Value states matrices $Q, K, V \in \R^{n \times d}$ be defined as Definition~\ref{def:attn}.
        \item $A  := \exp(Q K^\top / \sqrt{d}) \in \R^{n \times n}$.
        \item $D := \diag(A {\bf 1}_n)\in \R^{n \times n}$
        \item Let $\mathcal{T}_k$ be defined as Definition~\ref{def:T}.
        \item Let ${\sf topk}$ be defined as Definition~\ref{def:topk}.
        \item Let $\mathcal{S}_\epsilon$ be defined as Definition~\ref{def:S}, we omit $\mathcal{S}_{\epsilon} (D^{-1}_{i, i} A_{i, *})$ for $i \in [n]$ to $\mathcal{S}_{\epsilon, i}$.
        \item Define $A_{\rm spar} := \begin{bmatrix}
        {\sf topk}(A_{1, *}) & {\sf topk}(A_{2, *}) & \cdots & {\sf topk}(A_{n, *})
    \end{bmatrix}^\top$.
        \item $D_{\rm spar} := \diag(A_{\rm spar} {\bf 1}_n)\in \R^{n \times n}$.
        \item $\delta \in (0, 0.1)$.
        \item Let $R$ be defined as Definition~\ref{def:R}.
    \end{itemize}
    Then with a probability at least $1 - \delta$, we have
    \begin{itemize}
        \item {\bf Part 1.} Choosing $k = \Omega(n^C)$ for $C \in (0, 1)$, we have $C_{\rm error} \in (0, C)$:
        \begin{align*}
            \| D_{\rm spar}^{-1}A_{\rm spar} V - D^{-1}A V \|_\infty \leq o(n^{-C_{\rm error}})
        \end{align*}
        \item {\bf Part 2.} Choosing $k = o(\log(n))$ for $C \in (0, 1)$, we have $C_{\rm error} \in (0, C)$:
        \begin{align*}
            \| D_{\rm spar}^{-1}A_{\rm spar} V - D^{-1}A V \|_\infty \leq \Omega(n^{C_{\rm error}})
        \end{align*}
    \end{itemize}
\end{theorem}

\begin{proof}
    % Before we prove the result, we have
    % \begin{align}\label{eq:C_a}
    %     R = C_a \sqrt{\log(n)}
    % \end{align}
    % this step follows from Definition~\ref{def:R} and {\it General Conditions 5} of Definition~\ref{def:general_conditions}. Here, $C_a \in (0, 0.001)$ is a sufficiently small constant.

    % Then by Lemma~\ref{lem:upper_bound_error}, we can show that
    % \begin{align}\label{eq:C_b}
    %     \sqrt{\log(\poly(n)/\delta)} \leq \sqrt{C_b^2 \log(n)} = C_b\sqrt{\log(n)}
    % \end{align}
    % Here, $C_b \le 10$ is a constant.
    
    We have
    \begin{align}\label{eq:bounding_error}
        \| D_{\rm spar}^{-1}A_{\rm spar} - D^{-1}A \|_\infty 
        = & ~ \| D_{\rm spar}^{-1}A_{\rm spar} - D^{-1}A_{\rm spar} + D^{-1}A_{\rm spar} - D^{-1}A \|_\infty \notag \\
        \leq & ~ \| D_{\rm spar}^{-1}A_{\rm spar} - D^{-1}A_{\rm spar} \|_\infty + \| D^{-1}A_{\rm spar} - D^{-1}A \|_\infty \notag\\
        \leq & ~ (\frac{n-k}{nk} + \frac{1}{n}) \cdot \exp(O(R) \cdot \sqrt{\log(nd / \delta)}) \notag\\
        \leq & ~ \frac{1}{k} \cdot \exp(O(R) \cdot \sqrt{\log(nd / \delta)}) 
    \end{align}
    where the first step follows from simple algebras, the second step follows from triangle inequality, the third step follows from Part 1 and Part 4 of Lemma~\ref{lem:upper_bound_error}, the fourth step follows from simple algebras.

    {\bf Part 1.}
    Choosing $k = \Omega(n^C)$ for $C \in (0, 1)$, we have:
    \begin{align*}
        \| D_{\rm spar}^{-1}A_{\rm spar} - D^{-1}A \|_\infty 
        \leq & ~  \frac{1}{k} \cdot \exp(O(R) \cdot \sqrt{\log(nd / \delta)}) \\
        \leq & ~ o(n^{-C_{\rm error}})
    \end{align*}
    where the first step follows from Eq.~\eqref{eq:bounding_error}, the second step follows from $0 < C_{\rm error} < C$.

    {\bf Part 2.}
    Choosing $k = o(\log(n))$ for $C \in (0, 1)$, we have:
    \begin{align*}
        \| D_{\rm spar}^{-1}A_{\rm spar} - D^{-1}A \|_\infty 
        \leq & ~  \frac{1}{k} \cdot \exp(O(R) \cdot \sqrt{\log(nd / \delta)}) \\
        \leq & ~ \Omega(n^{C_{\rm error}})
    \end{align*}
    where the first step follows from Eq.~\eqref{eq:bounding_error}, the second step follows from $0 < C_{\rm error}$.
\end{proof}

\subsection{Main Theorem 4: Lower Bound on Error}

\begin{theorem}\label{thm:approx_failure}
    If the following conditions hold:
    \begin{itemize}
        \item Let Query, Key and Value states matrices $Q, K, V \in \R^{n \times d}$ be defined as Definition~\ref{def:attn}.
        \item $A  := \exp(Q K^\top / \sqrt{d}) \in \R^{n \times n}$.
        \item $D := \diag(A {\bf 1}_n)\in \R^{n \times n}$
        \item Let $\mathcal{T}_k$ be defined as Definition~\ref{def:T}.
        \item Let ${\sf topk}$ be defined as Definition~\ref{def:topk}.
        \item Let $\mathcal{S}_\epsilon$ be defined as Definition~\ref{def:S}, we omit $\mathcal{S}_{\epsilon} (D^{-1}_{i, i} A_{i, *})$ for $i \in [n]$ to $\mathcal{S}_{\epsilon, i}$.
        \item Define $A_{\rm spar} := \begin{bmatrix}
        {\sf topk}(A_{1, *}) & {\sf topk}(A_{2, *}) & \cdots & {\sf topk}(A_{n, *})
    \end{bmatrix}^\top$.
        \item $D_{\rm spar} := \diag(A_{\rm spar} {\bf 1}_n)\in \R^{n \times n}$.
        \item $\delta \in (0, 0.1)$.
        \item Let $R$ be defined as Definition~\ref{def:R}.
        \item Choosing $k = o(\log(n))$
    \end{itemize}
    Then with a probability at least $1 - \delta$, we have
    \begin{align*}
        \| D_{\rm spar}^{-1}A_{\rm spar} - D^{-1}A \|_F \ge O(1)
    \end{align*}
\end{theorem}

\begin{proof}
    We have:
    \begin{align*}
        \| D_{\rm spar}^{-1} A_{\rm spar} - D^{-1}A \|_F^2
        = & ~ \sum_{i_1=1}^n \sum_{i_2=1}^n (D_{{\rm spar}, i_1, i_1}^{-1} A_{{\rm spar}, i_1, i_2}  - D_{i_1, i_1}^{-1}A_{i_1, i_2} )^2 \\
        \ge & ~ \sum_{i_1=1}^n \sum_{i_2\in \mathcal{T}_k}  ( \frac{1}{k} \exp(-O(R) \cdot \sqrt{\log(\frac{nd}{\delta})}) - \frac{1}{n} \exp(O(R) \cdot \sqrt{\log(\frac{nd}{\delta})}) )^2 \\
        + & ~ \sum_{i_1=1}^n \sum_{i_2\in [n]/\mathcal{T}_k}  ( \frac{1}{n} \exp(-O(R) \cdot \sqrt{\log(\frac{nd}{\delta})}) )^2 \\
        \ge & ~ \sum_{i_1=1}^n \sum_{i_2\in \mathcal{T}_k}  ( O(\frac{1}{\sqrt{n \cdot o(\log(n))}}) )^2 \\
        \ge & ~ O(1)
    \end{align*}
    where the first step follows from the definition of Frobenius norm $\ell_F$, the second step follows from plugging $k = o(\sqrt{\log(n)})$ and we have:
    \begin{align*}
        \frac{\d}{\d n} \Big( \frac{1}{k} \exp(-O(R) \cdot \sqrt{\log(\frac{nd}{\delta})}) - \frac{1}{n} \exp(O(R) \cdot \sqrt{\log(\frac{nd}{\delta})}) \Big) \geq \frac{\d}{\d n} \frac{1}{\sqrt{n \cdot o(\log(n))}},
    \end{align*}
    and the last step follows from simple algebras.
\end{proof}

\subsection{Approximating Softmax Function}

\begin{lemma}\label{lem:upper_bound_error}
    If the following conditions hold:
    \begin{itemize}
        \item Let Query, Key and Value states matrices $Q, K, V \in \R^{n \times d}$ be defined as Definition~\ref{def:attn}.
        \item $A  := \exp(Q K^\top / \sqrt{d}) \in \R^{n \times n}$.
        \item $D := \diag(A {\bf 1}_n)\in \R^{n \times n}$
        \item Let $\mathcal{T}_k$ be defined as Definition~\ref{def:T}.
        \item Let ${\sf topk}$ be defined as Definition~\ref{def:topk}.
        \item Let $\mathcal{S}_\epsilon$ be defined as Definition~\ref{def:S}, we omit $\mathcal{S}_{\epsilon} (D^{-1}_{i, i} A_{i, *})$ for $i \in [n]$ to $\mathcal{S}_{\epsilon, i}$.
        \item Define $A_{\rm spar} := \begin{bmatrix}
        {\sf topk}(A_{1, *}) & {\sf topk}(A_{2, *}) & \cdots & {\sf topk}(A_{n, *})
    \end{bmatrix}^\top$.
        \item $D_{\rm spar} := \diag(A_{\rm spar} {\bf 1}_n)\in \R^{n \times n}$.
        \item $\delta \in (0, 0.1)$.
        \item Let $R \ge 0$ be defined as Definition~\ref{def:R}.
    \end{itemize}
    Then with a probability at least $1 - \delta$, we have
    \begin{itemize}
        \item Part 1.
        \begin{align*}
            \| D^{-1}A_{\rm spar} - D^{-1}A \|_\infty \leq \frac{1}{n} \cdot \exp( O(R) \cdot \sqrt{\log(nd / \delta)})
        \end{align*}
        \item Part 2.
        \begin{align*}
            \| D_{\rm spar} - D \|_\infty \leq (n-k) \cdot \exp( O(R) \cdot \sqrt{\log(nd / \delta)})
        \end{align*}
        \item Part 3.
        \begin{align*}
            \| D_{\rm spar}^{-1} - D^{-1} \|_\infty \leq \frac{n-k}{nk} \cdot \exp(O(R) \cdot \sqrt{\log(nd / \delta)})
        \end{align*}
        \item Part 4.
        \begin{align*}
            \| D_{\rm spar}^{-1}A_{\rm spar} - D^{-1}A_{\rm spar} \|_\infty \leq \frac{n-k}{nk} \cdot \exp( O(R) \cdot \sqrt{\log(nd / \delta)})
        \end{align*}
    \end{itemize}
\end{lemma}

\begin{proof}
    Before we begin the proof, we construct a toolkit as follows:

    For $x_1 \leq \exp(\sqrt{\log(a/\delta_1)})$ and $x_2 \leq \exp(\sqrt{\log(a/\delta_2)})$, we have
    \begin{align}\label{eq:combine_exp_sqrt_log}
        x_1 x_2 
        \leq & ~ \exp(\sqrt{\log(a/\delta_1)}) \cdot \exp(\sqrt{\log(b/\delta_2)}) \notag \\
        \leq & ~ \exp(\sqrt{\log(a/\delta_1)} + \sqrt{\log(b/\delta_2)}) \notag \\
        \leq & ~ \exp(C\sqrt{\log(a/\delta_1) + \log(b/\delta_2)}) \notag \\
        \leq & ~ \exp(C\sqrt{\log(ab/\delta)})
    \end{align}
    where these steps follow from simple algebras, Fact~\ref{fac:exp_sqrt_log} and choose $\delta_1 = \delta_2 = \delta/2$.
    
    {\bf Proof of Part 1.}
    This proof follows from Theorem~\ref{thm:main_result} and $n - k \leq n$.

    {\bf Proof of Part 2.}
    We have
    \begin{align*}
        \| D_{\rm spar} - D \|_\infty
        = & ~ \| A_{\rm spar} {\bf 1}_n - A {\bf 1}_n \|_\infty \\
        = & ~  \| D \circ (D^{-1} A_{\rm spar} {\bf 1}_n - D^{-1} A {\bf 1}_n ) \|_\infty \\
        \leq & ~ \| D\|_\infty \cdot \| D^{-1}A_{\rm spar} {\bf 1}_n - D^{-1}A {\bf 1}_n \|_\infty \\
        \leq & ~ \| D\|_\infty \cdot \frac{n-k}{n} \cdot \exp(O(R) \cdot \sqrt{\log(nd / \delta)}) \\
        \leq & ~ (n-k) \cdot \exp(O(R) \cdot \sqrt{\log(nd / \delta)}) \cdot \exp(O(R) \cdot \sqrt{\log(nd/\delta)}) \\
        \leq & ~ (n-k) \cdot \exp(O(R) \cdot \sqrt{\log(nd / \delta)}) 
    \end{align*}
    where the first step follows from the definitions of $D_{\rm spar}$ and $D$, the second step follows from simple algebras, the third step follows from Cauchy-Schwarz inequality, the fourth step follows from Part 1 of this Lemma and the definition of $A_{\rm spar}$, the fifth step follows from Part 4 of Lemma~\ref{lem:bound_tool}, the sixth step follows from Eq.~\eqref{eq:combine_exp_sqrt_log}.

    {\bf Proof of Part 3.}
    We have
    \begin{align*}
        \| D_{\rm spar}^{-1} - D^{-1} \|_\infty
        = & ~ \| D_{\rm spar}^{-1} \|_\infty \cdot \| D^{-1} \|_\infty \cdot \| D_{\rm spar} - D\|_\infty \\
        \leq & ~ \| D_{\rm spar}^{-1} \|_\infty \cdot \| D^{-1} \|_\infty \cdot  (n-k) \cdot \exp(O(R) \cdot \sqrt{\log(nd / \delta)}) \\
        \leq &  ~ \frac{n-k}{nk} \cdot \exp(O(R) \cdot \sqrt{\log(nd / \delta)}) \cdot \exp(O(R) \cdot \sqrt{\log(nd/\delta)}) \\
        \leq & ~ \frac{n-k}{nk} \cdot \exp(O(R) \cdot \sqrt{\log(nd / \delta)})
    \end{align*}
    where the first step follows from simple algebras, the second step follows from Part 2 of this Lemma, the third step follows from Part 5 and Part 6 of Lemma~\ref{lem:bound_tool}, the last step follows from Eq.~\eqref{eq:combine_exp_sqrt_log}.

    {\bf Proof of Part 4.}
    This proof follows from combining Part 2 of Lemma~\ref{lem:bound_tool} and Part 3 of this Lemma.
\end{proof}

\subsection{Helpful Bound Toolkit}

\begin{lemma}\label{lem:bound_tool}
    If the following conditions hold:
    \begin{itemize}
        \item Let Query, Key and Value states matrices $Q, K, V \in \R^{n \times d}$ be defined as Definition~\ref{def:attn}.
        \item $A  := \exp(Q K^\top / \sqrt{d}) \in \R^{n \times n}$.
        \item $D := \diag(A {\bf 1}_n)\in \R^{n \times n}$
        \item Let $\mathcal{T}_k$ be defined as Definition~\ref{def:T}.
        \item Let ${\sf topk}$ be defined as Definition~\ref{def:topk}.
        \item Let $\mathcal{S}_\epsilon$ be defined as Definition~\ref{def:S}, we omit $\mathcal{S}_{\epsilon} (D^{-1}_{i, i} A_{i, *})$ for $i \in [n]$ to $\mathcal{S}_{\epsilon, i}$.
        \item Define $A_{\rm spar} := \begin{bmatrix}
        {\sf topk}(A_{1, *}) & {\sf topk}(A_{2, *}) & \cdots & {\sf topk}(A_{n, *})
    \end{bmatrix}^\top$.
        \item $D_{\rm spar} := \diag(A_{\rm spar} {\bf 1}_n)\in \R^{n \times n}$.
        \item $\delta \in (0, 0.1)$.
        \item Let $R \ge 0$ be defined as Definition~\ref{def:R}.
    \end{itemize}
    Then with a probability at least $1 -\delta$, we have
    \begin{itemize}
        \item Part 1. $\exp(-O(R) \cdot \sqrt{\log(\frac{n-k}{\delta}d)}) \leq A_{i_1, i_2} \leq \exp(O(R) \cdot \sqrt{\log(\frac{n-k}{\delta}d)})$, $\forall i_1 \in [n], i_2 \in {\cal S}_\epsilon$
        \item Part 2. $\exp(-O(R) \cdot \sqrt{\log(\frac{n}{\delta}d)}) \leq A_{i_1, i_2} \leq \exp(O(R) \cdot \sqrt{\log(\frac{n}{\delta}d)})$, $\forall i_1, i_2 \in [n]$
        \item Part 3. $k \exp(-O(R) \cdot \sqrt{\log(\frac{n}{\delta}d)}) \leq \sum_{i_2 \in {\cal T}_k} A_{i_1, i_2} \leq k \exp(O(R) \cdot \sqrt{\log(\frac{n}{\delta}d)})$, $\forall i_1 \in [n]$
        \item Part 4. $ n \exp(-O(R) \cdot \sqrt{\log(\frac{n}{\delta}d)}) \leq D_{i_1, i_1} \leq n \exp(O(R) \cdot \sqrt{\log(\frac{n}{\delta}d)})$, $\forall i_1 \in [n]$
        \item Part 5. $\frac{1}{k} \exp(-O(R) \cdot \sqrt{\log(\frac{n}{\delta}d)}) \leq \Big( \sum_{i_2 \in {\cal T}_k} A_{i_1, i_2} \Big)^{-1} \leq \frac{1}{k} \exp(O(R) \cdot \sqrt{\log(\frac{n}{\delta}d)})$, $\forall i_1 \in [n]$
        \item Part 6. $ \frac{1}{n} \exp(-O(R) \cdot \sqrt{\log(\frac{n}{\delta}d)}) \leq D_{i_1, i_1}^{-1} \leq \frac{1}{n} \exp(O(R) \cdot \sqrt{\log(\frac{n}{\delta}d)})$, $\forall i_1 \in [n]$
    \end{itemize}
\end{lemma}

\begin{proof}
    {\bf Proof of Part 1.}

    This proof follows from Eq.~\eqref{eq:bound_A}.

    {\bf Proof of Part 2.}

    This proof follows from Part 1 and Part 2 of Lemma~\ref{lem:bound_D}.

    {\bf Proof of Part 3.}

    This proof follows from Part 1 of this Lemma.

    {\bf Proof of Part 4}.

    This proof follows from Part 2 of Lemma~\ref{lem:bound_D}.

    {\bf Proof of Part 5.}

    This proof follows from Part 3 of this Lemma.

    {\bf Proof of Part 6.}
    This proof follows from Part 4 of this Lemma.
\end{proof}